%% file: main.tex
\documentclass[10pt,twocolumn,letterpaper]{article}
\usepackage[accsupp]{axessibility}
\usepackage{iccv}
\usepackage{times}
\usepackage{epsfig}
\usepackage{graphicx}
\usepackage{amsmath}
\usepackage{amssymb}
\usepackage{float}
\usepackage{capt-of}
\usepackage{cuted}
\usepackage{comment}
\usepackage{xcolor}
\usepackage{color}
\usepackage[pagebackref=true,breaklinks=true,letterpaper=true,colorlinks,bookmarks=false]{hyperref}
\usepackage[capitalise]{cleveref}
\usepackage[linesnumbered,ruled,vlined]{algorithm2e}

\hypersetup{urlcolor=pink}
\definecolor{pink}{HTML}{EC008C}

\hypersetup{citecolor=orange}
\hypersetup{linkcolor=orange}
\hypersetup{linkcolor=orange}
\hypersetup{urlcolor=pink}

\usepackage[pagebackref=true,breaklinks=true,letterpaper=true,colorlinks,bookmarks=false]{hyperref}

\iccvfinalcopy % *** Uncomment this line for the final submission

\def\iccvPaperID{4211} % *** Enter the ICCV Paper ID here

\ificcvfinal\fi

\begin{document}

%%%%%%%%% TITLE
\title{Texture Generation on 3D Meshes with Point-UV Diffusion}

\author{
Xin Yu$^{1}$ 
% \\ {\tt\small yuxin27g@gmail.com}
\and 
Peng Dai$^{1}$ 
% \\ {\tt\small daipeng@eee.hku.hk}
\and 
Wenbo Li$^2$ 
%\\ {\tt\small wenboli@cse.cuhk.edu.hk}
\and
Lan Ma$^3$ 
%\\ {\tt\small rubyma@tcl.com}
\and
Zhengzhe Liu$^{2\dagger}$ 
%\\ {\tt\small zzliu@cse.cuhk.edu.hk}
\and
Xiaojuan Qi$^{1\dagger}$ 
%\\ {\tt\small xjqi@eee.hku.hk}
\and
\\
$^1$The University of Hong Kong,~\,
$^2$The Chinese University of Hong Kong,~\,
$^3$TCL Corporate Research\\
\tt\small \{yuxin,daipeng,xjqi\}@eee.hku.hk \qquad
  \tt \{wenboli,zzliu\}@cse.cuhk.edu.hk \qquad rubyma@tcl.com \\
}

\ificcvfinal\thispagestyle{empty}\fi

%%%%%%%%% ABSTRACT
\input{./main_body.tex}

{\small
\bibliographystyle{ieee_fullname}
\bibliography{egbib}
}

\clearpage
\centerline{\Large \textbf{Appendix}}
\renewcommand*{\thesection}{\Alph{section}}
\setcounter{section}{0}
\vspace{4pt}

\input{./supplement}

\end{document}

%% file: main_body.tex
\twocolumn[
\maketitle
\vspace{-2em}
\input{my_figures/00_teaser.tex}
%\medbreak
\vspace{1.0em}
]

\begin{abstract}
\vspace{-1.2em}
\input{00_abstract}
\end{abstract}

\input{01_intro}

\input{my_figures/02_compare.tex}
\input{my_figures/01_uvmap.tex}
\input{02_relatedworks}
\input{my_figures/03_framework.tex}
\input{03_preliminaries}
\input{04_method}

\input{05_experiments}
\input{06_conclusion.tex}

%% file: my_figures/00_teaser.tex
\centering
\includegraphics[width=1.0\linewidth]{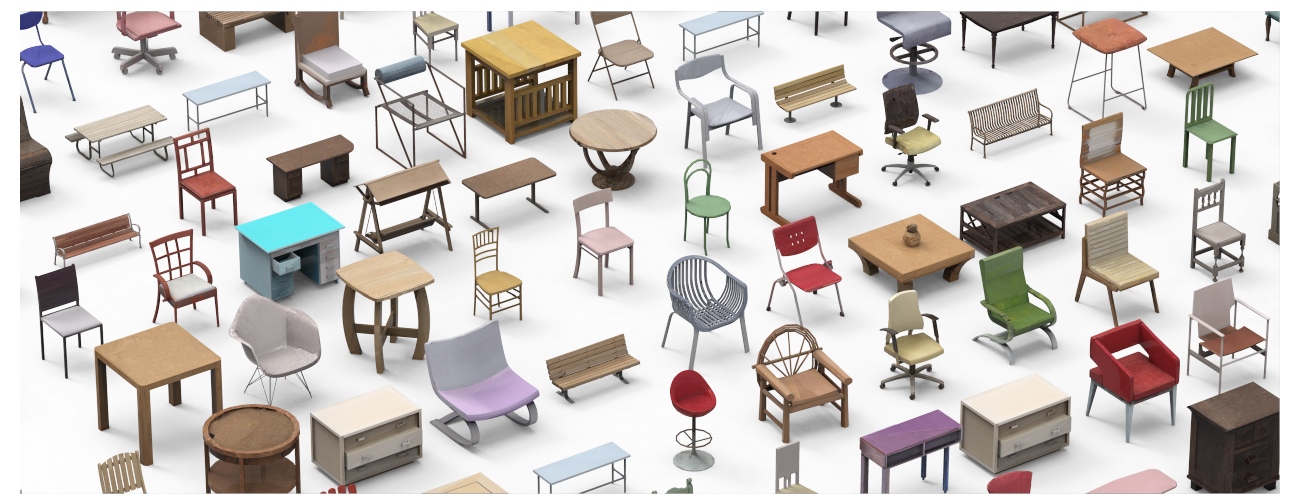}
%\vspace{-10pt}
\captionof{figure}{\textbf{A gallery of generated results by our Point-UV diffusion.} Our method is capable of processing meshes of any genus, generating diversified, geometry-compatible, and high-fidelity textures. 
%\vspace{10pt}
\label{fig:teasing}} 

%% file: 00_abstract.tex
%\vspace{-0.11in} 
In this work, we focus on synthesizing high-quality textures on
3D meshes. 
We present Point-UV diffusion, a coarse-to-fine pipeline that marries the denoising diffusion model with UV mapping to generate 3D consistent and high-quality texture images in UV space. 
We start with introducing a point diffusion model to synthesize low-frequency texture components with our tailored style guidance to tackle the biased color distribution.
The derived coarse texture offers global consistency and 
serves as a condition for the subsequent UV diffusion stage, aiding in regularizing the model to generate a 3D consistent UV texture image.
Then, a UV diffusion model with hybrid conditions is developed 
 to enhance the texture fidelity in the 2D UV space. Our method can process meshes of any genus, generating
diversified, geometry-compatible, and high-fidelity textures. Code is available at  \href{https://cvmi-lab.github.io/Point-UV-Diffusion}{https://cvmi-lab.github.io/Point-UV-Diffusion}. 

\let\thefootnote\relax\footnotetext{$^{\dagger}$: Corresponding authors}

%% file: 01_intro.tex
\section{Introduction}

%Texture is a fundamental element in 3D meshes.
Texturing 3D meshes is a fundamental task in computer vision and graphics. It enhances the visual richness of 3D objects, thereby facilitating their application in various fields such as video games, 3D movies, and AR/VR technologies. However, generating high-quality textures can be daunting and time-consuming, often requiring specialized knowledge and resources. As such, there is a pressing need for an efficient approach to automatically create high-quality textures on 3D meshes.

Despite the substantial progress the community has made in  2D image synthesis and 3D shape generation using GANs~\cite{gulrajani2017improved,karras2019style,wang2021cycle,gao2022get3d} or diffusion models~\cite{ramesh2022hierarchical,saharia2022photorealistic,rombach2022high,hui2022neural}, crafting realistic textures on mesh surfaces remains challenging. 
One major difficulty stems from the need for suitable 3D representations for texture synthesis.
Early approaches investigate the use of voxels~\cite{chaudhuri2021semi,yin20213dstylenet,chen2019text2shape} or point clouds~\cite{gao2020learning} and synthesize point/voxel colors.  
However, they can only afford to synthesize low-resolution results with low-fidelity textures due to memory and model complexity constraints. 
In response, Texture Fields~\cite{oechsle2019texture} adopts an implicit representation with the potential to synthesize high-resolution textures, but still could not yield satisfactory results as shown in Figure~\ref{fig:generated} (a): over-smoothed results. 
Most recently, Siddiqui \etal \cite{siddiqui2022texturify}  propose to parameterize the shape as tetrahedral meshes and introduce tetrahedral mesh convolution to enhance local details. 
Albeit improving results, tetrahedral parameterization inevitably destroys geometric details of the input mesh and thus cannot faithfully preserve the original structure. As shown in Figure~\ref{fig:generated} (b), the delicate structures of the chair's back are absent. Moreover, the generative models utilized in these methods are limited to GANs~\cite{oechsle2019texture,siddiqui2022texturify} and VAEs~\cite{oechsle2019texture,gao2021tm}. The more advanced diffusion model, which could potentially open up new avenues for high-quality texture generation, remains insufficiently explored.

In this paper, we delve into a novel texture representation based on UV maps and investigate the advanced diffusion model for texture generation. The 2D nature of the UV map enables it to circumvent the cost of high-resolution point/voxel representations. Besides, the UV map is compatible with arbitrary mesh topologies, thereby preserving the original geometric structures.
However, while promising, direct integration of the UV map representation with a 2D diffusion model presents challenges in synthesizing seamless textures, leading to severe artifacts, as shown in Figure~\ref{fig:generated} (c). This occurs because the UV mapping process fragments the continuous texture on the 3D surface into isolated patches on the 2D UV plane (see Figure~\ref{fig:uvmap}). 

To this end, we introduce Point-UV diffusion, a two-stage coarse-to-fine framework consisting of point diffusion and UV diffusion.  
Specifically, we initially design a point diffusion model to generate color for sampled points that act as low-frequency texture components. This model is equipped with a style guidance mechanism that alleviates the impact of biased color distributions in the dataset and facilitates diversity during inference.
Next, we project these colorized points onto the 2D UV space with 3D coordinate interpolation, thereby generating a coarse texture image that maintains 3D consistency and continuity. Given the coarse textured image, we develop a UV diffusion model with elaborately designed hybrid conditions to improve the quality of the textures (see Figure \ref{fig:generated} (ours) and Figure \ref{fig:teasing}).

In short, our contributions are as follows: 1) We propose a  new framework for texture generation for given meshes. Our representation can handle meshes with arbitrary topology and is able to faithfully preserve geometric structures. 2) To the best of our knowledge, we are the first to train a diffusion model specifically for mesh texture generation. Our coarse-to-fine framework allows us to enjoy the efficiency of 2D representation while enhancing 3D consistency. 3) We compare our approach with multiple methods in unconditional generation and achieve state-of-the-art results. Furthermore, we demonstrate that our method can be easily extended to scenarios with text-conditioning and image-conditioning.

%% file: my_figures/02_compare.tex
\begin{figure}[t]
\raggedright
    \includegraphics[trim={0cm 0cm 0cm 0cm},clip,width=\columnwidth]{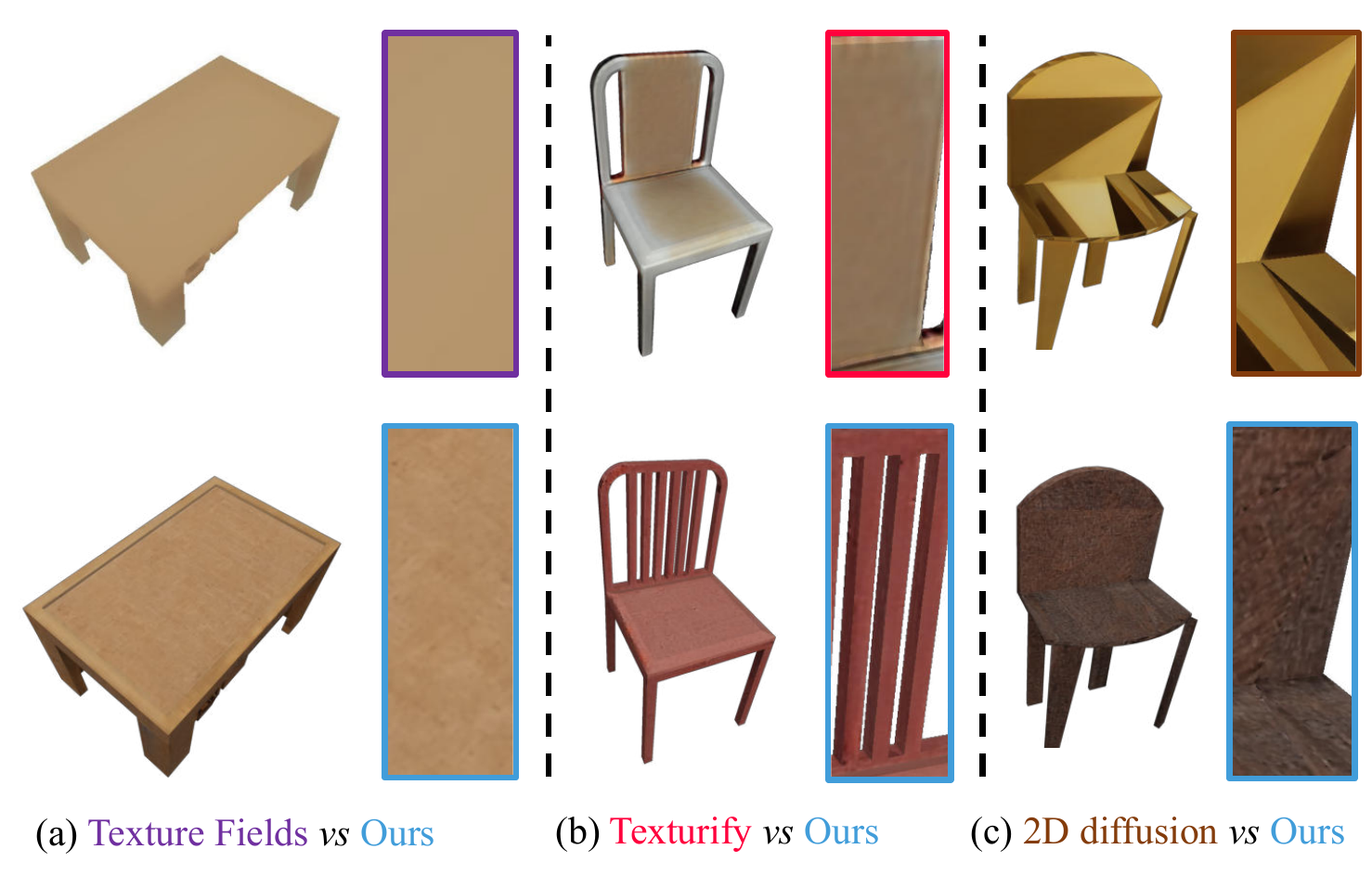}
    \caption{\textbf{Comparisons with different methods.} Our generative results  (a) possess high-quality details, (b) faithfully preserve the mesh structure, and (c) are better consistent with the given shape, compared with Texture Fields~\cite{oechsle2019texture}, Texturify~\cite{siddiqui2022texturify} and 2D diffusion~\cite{ho2020denoising}, respectively.
    }
    \label{fig:generated}
\end{figure}

%% file: my_figures/01_uvmap.tex
\begin{figure}[t]
    \includegraphics[trim={0cm 0cm 0cm 0cm},clip,width=\columnwidth]{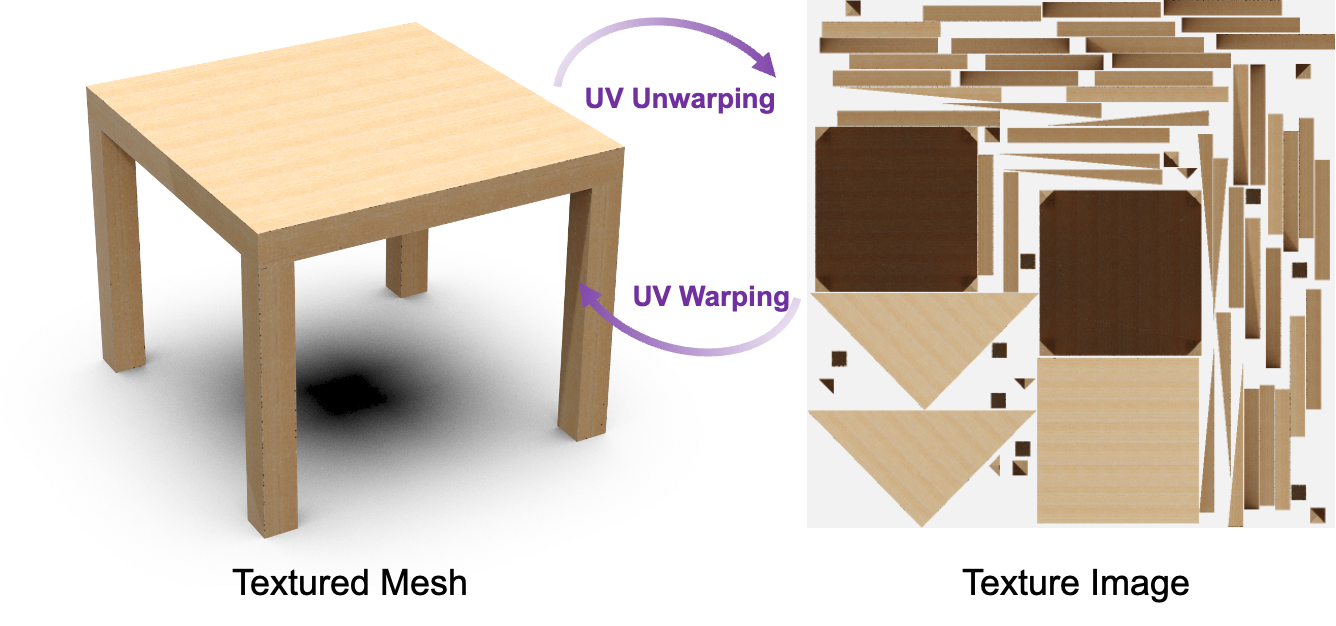}
    \caption{\textbf{Illustration of UV mapping process.} It establishes connections between the 2D texture map and the surface appearance of 3D shape.}
    \label{fig:uvmap}
\end{figure}

%% file: 02_relatedworks.tex
\section{Related Work}

\paragraph{Texture generation.}
Early works~\cite{sun2018im2avatar, tulsiani2017multi,chen2019text2shape} propose using voxels to represent colors. However, due to a cubic increase in memory usage and computational cost with rising resolution, these methods produce only coarse, low-resolution textures. For meshes, recent work in texture generation~\cite{siddiqui2022texturify} predicts the color for each face, but the representation capacity remains largely constrained by the mesh's resolution. Another set of works~\cite{pavllo2020convolutional,tulsiani2020implicit,mohammad2022clip,gao2021tm} leverage UV mapping, but they require either spherical parameterization or part segmentation and, therefore, are confined to handling low-genus shapes. Other works~\cite{portenier2020gramgan, henzler2020learning} rely on image exemplars, while our approach focuses on unconditional generation. Recently, implicit functions attract increasing attention for generative tasks~\cite{gao2022get3d, poole2022dreamfusion, skorokhodov2022epigraf, chan2021pi, deng2022gram, schwarz2020graf, oechsle2019texture,liu2022iss}. Most of this research centers on generating 3D-aware images~\cite{skorokhodov2022epigraf, chan2021pi, deng2022gram, schwarz2020graf}, rather than synthesizing textures for given 3D meshes. Texture Fields~\cite{oechsle2019texture} represents the most similar work to our task, which, however, tends to produce over-smoothed results. 

Some methods focus on test-time optimization 
for generation tasks. For instance, Text2Mesh~\cite{michel2022text2mesh} 
employs CLIP~\cite{radford2021learning} to design 
a loss function. This function stylizes a 3D 
mesh to align with a target text prompt, 
predicting both the color and displacement 
for each mesh vertex. More recently, DreamFusion~\cite{poole2022dreamfusion} and Magic3D~\cite{lin2023magic3d} leverage pre-trained stable diffusion~\cite{rombach2022high} for score distillation, creating either NeRF~\cite{mildenhall2021nerf} 
or a 3D mesh. However, these approaches 
often demand extensive optimization time 
or necessitate alterations to the existing 
mesh geometry~\cite{michel2022text2mesh}. In this paper, 
our objective is to generate textures for specified 
full-3D meshes with arbitrary topology. This goal 
entails two fundamental requirements: 1) preserving 
the original mesh's geometric structure; 2) producing 
a 3D texture representation exportable as 
either vertex color or a uv-texture-image, 
instead of merely generating/rendering multi-view 
or 3D-aware images.

\paragraph{Diffusion models for 3D generation.}
In addition to 2D image generation, 
diffusion models recently gain significant attention 
in 3D generation, leading to many 
related works~\cite{zhou20213d, hu2023neural, hui2022neural, 
nichol2022point, poole2022dreamfusion, zeng2022lion, anciukevivcius2022renderdiffusion}. 
For instance, Zhou et al. \cite{zhou20213d} introduce 
point-voxel diffusion for point cloud generation 
and completion. Hui et al. \cite{hui2022neural} 
and Hu et al. \cite{hu2023neural} suggest 
a compact wavelet domain representation for shapes, 
enabling higher-quality shape creation via diffusion 
models. Nichol et al.~\cite{nichol2022point} present Point-E, 
a system that uses a text-conditional strategy 
to produce colored 3D point clouds. This approach 
enables an efficient synthesis of intricate 3D shapes 
from textual prompts. Yet, this system yields 
lower-resolution point clouds, often missing detailed 
shapes and textures. In this work, we develop a brand new 3D diffusion model for texture image synthesis, which allows high-fidelity and 3D-consistent texture generation.

%% file: my_figures/03_framework.tex
\begin{figure*}[t]
    \includegraphics[width=\textwidth]{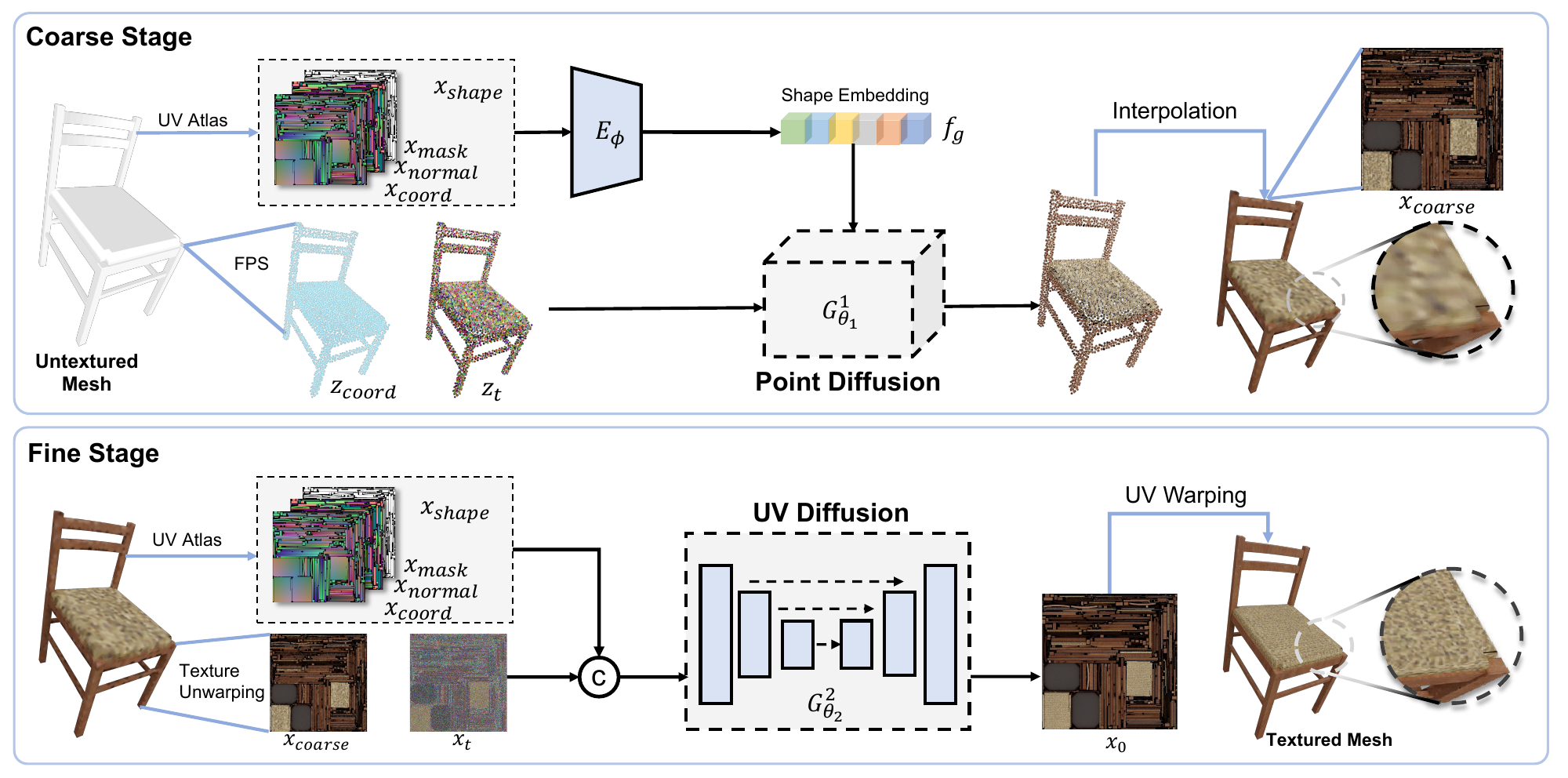}
    \caption{\textbf{The overview of our Point-UV-diffusion framework.} (Top) The coarse stage first samples a point cloud on the mesh surface, and then predicts the color for each point using a 3D diffusion model conditioned on shape features including surface normal, mask, and coordinates. Then the points are mapped to the 2D UV map and the remaining uncolored points are filled up by tri-linear interpolation based on the 3D coordinates. (Bottom) The fine stage predicts high-quality textures with a 2D diffusion model conditioned on the shape attributes and the coarse texture image. } 
    \label{fig:framework}
\end{figure*}

%% file: 03_preliminaries.tex
\section{Preliminaries}
\label{sec:preliminaries}
    
\subsection{Denoising diffusion model.}
Diffusion model \cite{sohl2015deep, ho2020denoising} is a kind of likelihood-based generative model that has gained significant attention recently. It learns the data distribution $q(x_0)$ by progressive denoising from a prior Gaussian distribution. 
Given a sample from the data distribution $x_0 \sim q(x_0)$, a fixed forward process $q\left(\boldsymbol{x}_{1: T} \mid \boldsymbol{x}_0\right)=\prod_{t=1}^T q\left(\boldsymbol{x}_t \mid \boldsymbol{x}_{t-1}\right)$ is used to perturb the data with Gaussian kernels $q\left(\mathbf{x}_t \mid \mathbf{x}_{t-1}\right):=\mathcal{N}\left(\sqrt{1-\beta_t} \mathbf{x}_{t-1}, \beta_t \mathbf{I}\right)$, producing increasingly noisy latent variables $\{x_{1}, x_{2}, ..., x_{T}\}$.
Then, a parameterized Markov process $p_\theta\left(\boldsymbol{x}_{0: T}\right)=p\left(\boldsymbol{x}_T\right) \prod_{t=1}^T p_\theta\left(\boldsymbol{x}_{t-1} \mid \boldsymbol{x}_t\right)$ with transition kernel $p_\theta\left(\mathbf{x}_{t-1} \mid \mathbf{x}_t\right):=\mathcal{N}\left(\mu_\theta\left(\mathbf{x}_t, t\right), \sigma_t^2 \mathbf{I}\right)$ is optimized through maximizing a variational lower bound of log data likelihood, which essentially targets to match the joint distribution $q\left(\boldsymbol{x}_{0: T}\right)$: 
\begin{equation}
   \mathrm{E}_{q\left(\mathbf{x}_0\right)}\left[\log p_\theta\left(\mathbf{x}_0\right)\right] \geq \mathrm{E}_{q\left(\mathbf{x}_{0: T}\right)}\left[\log \frac{p_\theta\left(\mathbf{x}_{0: T}\right)}{q\left(\mathbf{x}_{1: T} \mid \mathbf{x}_0\right)}\right].
\end{equation}

After training, novel samples can then be generated via iterative sampling from $p_\theta\left(\boldsymbol{x}_{t-1} \mid \boldsymbol{x}_t\right)$ following:
$$
\mathbf{x}_{t-1}=\mu_\theta\left(\mathbf{x}_t, t\right)+\sqrt{\beta_t} \mathbf{z}, \mathbf{z} \sim \mathcal{N}(0, \mathbf{I}).
$$

Diffusion models have shown impressive capabilities in generating high-quality and diverse content. This paper proposes a texture generation framework based on diffusion model.

\subsection{UV mapping and challenges.}
\label{sec:challenge}
UV mapping is a method of surface parameterization that translates a 3D surface into a 2D image, effectively creating a 2D coordinate system known as a UV map for a polygonal mesh $S$. This is achieved by explicitly assigning UV coordinates to each vertex of the mesh. Further, an arbitrary surface coordinate on the mesh can be mapped to its 2D coordinate through barycentric interpolation:  
    \begin{equation}
        (u, v)=f(p) \quad f: S \rightarrow \Omega,
    \end{equation}
where $f$ represents the UV mapping process.

As illustrated in Figure~\ref{fig:uvmap} ``UV warping'', we can apply textures to the mesh surface by generating a high-resolution texture image in the UV space.
Besides texture, we can also create this kind of 2D map for surface normals and point coordinates.
Our objective is to generate a high-quality 2D UV texture image of the given mesh. 
However, the mapping process requires cutting the continuous texture on the 3D shape into a series of individual patches in the 2D UV plane, as depicted in Figure~\ref{fig:uvmap}. This fragmentation makes it challenging for the generative model to directly learn the 3D adjacency relationships of the patches within the 2D texture UV map. Consequently, this can lead to discontinuity and inconsistency issues when the generated texture map is applied back to the 3D mesh surface.
As shown in Figure~\ref{fig:generated} ``2D Diffusion'', the diffusion model generates inconsistent textures and suffers from discontinuity issues. 

%% file: 04_method.tex
\section{Method}
\label{sec:method}
The challenge mentioned in Section~\ref{sec:challenge} motivates us to 
propose a coarse-to-fine framework for 
texture image synthesis, namely Point-UV diffusion, 
illustrated in Figure~\ref{fig:framework}. 
To start, we design a 3D point diffusion model to colorize a set of sampled points on the mesh surface, as shown in Figure~\ref{fig:framework} (Top). This stage leverages the 3D topology for predicting low-frequency colors on the mesh, without being affected by the discontinuity of the UV map.  
Based on the color components generated by the coarse stage, we then establish a 
2D diffusion model in the UV space. This enhances the fidelity of the 
generated texture, as depicted in Figure~\ref{fig:framework} (Bottom).

\subsection{Coarse Stage: Point Diffusion} 
\label{sec:coarse}
In the coarse stage, we begin by 
executing farthest point sampling (FPS) on 
the mesh surface, deriving a point 
set consisting of $K$ points. 
These 
points are defined by their coordinates $\mathbf{z}_{\text{coord}}$ 
and colors $\mathbf{z}_{0}$. 
During training, a forward diffusion process 
degrades the clean colors $\mathbf{z}_{0}$, transforming them 
into a noisy state $\mathbf{z}_{t}$. The noise level is dictated by the time 
step $t$, where $t \in \{0, 1, ..., T\}$.

Our network is trained to reverse this diffusion process, aiming to denoise 
$\mathbf{z}_{t}$ back to its original clean colors. 
This denoising network is informed by 
three pre-computed maps: a coordinate map 
$\mathbf{x}_{\text{coord}}$, a normal map $\mathbf{x}_{\text{normal}}$, and a 
mask map $\mathbf{x}_{\text{mask}}$ (details in Section~\ref{sec:challenge}).
Unless stated otherwise, these conditions concatenate along 
the channel dimension, culminating in what 
we term the shape map $\mathbf{x}_{\text{shape}}$:
\begin{equation}
\mathbf{x}_{\text{shape}}=\left(\left[\mathbf{x}_{\text{normal}}, \mathbf{x}_{\text{mask}}, \mathbf{x}_{\text{coord}}\right]\right).
\end{equation}
Our network architecture is constructed 
upon PVCNN~\cite{liu2019point}, drawing similarities 
to point-voxel diffusion~\cite{zhou20213d}. However, 
we introduce slight modifications to amplify 
the integration of global shape information, contrasting 
with the approach in~\cite{zhou20213d} which primarily 
relies on point coordinates.
Initially, we employ a lightweight shape encoder $E_{\phi}$ to extract a global shape embedding $f_g$ from $\mathbf{x}_{\text{shape}}$. This embedding is subsequently fed into the 3D network $G^{1}_{\theta_1}$ along with $\mathbf{z}_t$, $\mathbf{z}_{\text{coord}}$, and $t$ to predict the color $\hat{\mathbf{z}}_0$:
\begin{equation}
\begin{aligned}
f_g&=E_{\phi}\left(\mathbf{x}_{\text{shape}}\right),\\    
\hat{\mathbf{z}}_0&=G^{1}_{\theta_1}\left(\left[\mathbf{z}_{\text{coord}}, \mathbf{z}_{t}, f_g, t\right]\right).
\end{aligned}
\end{equation}

\paragraph{Style guidance.}
We observe that the synthesized colors are largely influenced by the predominant colors within the dataset (for instance, textures in the ShapeNet ``chair" category are typically white, pure magenta, or wood-colored), leading to a lack of diversity in the outputs. 
To address this bias and promote color diversity, 
we introduce a style guidance mechanism.
This is achieved by flattening each $\mathbf{z}_0$ into a 
unidimensional vector, followed by employing PCA to extract 
the principal component coefficients, thus reducing the dimensionality 
of $\mathbf{z}_0$. Subsequently, K-means clustering is utilized 
to assign a style label to each $\mathbf{z}_0$. As depicted in Figure~\ref{fig:cluster}, shapes within a particular cluster exhibit similar color styles.
During training, we provide the network with an additional style label $z_{\text{style}}$ as a condition, which is referred to as style guidance:
\begin{equation}
\hat{\mathbf{z}}_0=G^{1}_{\theta_1}\left(\left[\mathbf{z}_{\text{coord}}, \mathbf{z}_{t}, f_g, t, z_{\text{style}}\right]\right).
\end{equation}
In this way, we can guide the network to predict the desired color during inference by providing a certain style condition, thereby alleviating the biased color issue.

\paragraph{Coarse texture image.}
With the colored point clouds, we then project them onto the 2D UV space based on the pre-calculated UV mapping. Following this, we perform KNN interpolation based on the 3D coordinates to assign colors to the remaining uncolored pixels, thereby generating a \textit{coarse texture image} $\mathbf{x}_{\text{coarse}}$ for the given mesh. This coarse texture image offers a coherent base color initialization and functions as a conditional element to guide the high-fidelity texture generation in the second stage, helping to avoid discontinuity caused by UV mapping.

\vspace{2pt}
\subsection{Fine Stage: UV Diffusion} 

To further refine the coarse texture image, we design a fine stage using 2D diffusion in the UV space, as depicted in Figure~\ref{fig:framework} (Bottom). In addition to the conditions used in the coarse stage, we incorporate the coarse texture image $\mathbf{x}_{\text{coarse}}$ as an additional condition.
We apply a 2D U-Net $G^{2}_{\theta_2}$ combined with self-attention modules to learn the high-quality texture image $\hat{\mathbf{x}}_0$, from a noisy texture image $\mathbf{x}_t$:
\begin{equation}
\hat{\mathbf{x}}_0=G^{2}_{\theta_2}\left(\left[\mathbf{x}_{\text{shape}}, \mathbf{x}_{t}, \mathbf{x}_{\text{coarse}}, t\right]\right).
\end{equation} 

\paragraph{Hybrid condition.}
During training, we employ FPS on the ground-truth texture image, $\mathbf{x}_0$,
followed by interpolation to simulate the coarse texture image,
$\mathbf{x}_{\text{coarse}}$.
However, in joint cascaded testing involving both stages,
we note that the output quality of $\hat{\mathbf{x}}_{\text{coarse}}$
from the first stage doesn't always align flawlessly with the quality
of $\mathbf{x}_{\text{coarse}}$. Such mismatches can influence the
performance of the subsequent stage. 
To address this,
we introduce a hybrid conditioning method aimed at narrowing this discrepancy. 
Firstly, we create a smooth texture image $\mathbf{x}_{\text{smooth}}$ (see Figure~\ref{fig:coarse}), inspired by the blur augmentation described in~\cite{ho2022cascaded} for cascaded diffusion models. In particular, we segment the mask map into
multiple discrete regions using four-connectivity detection,
then perform average color pooling within each region based on
its connectivity. After this procedure, $\mathbf{x}_{\text{smooth}}$
maintains merely the regional color,
ensuring more consistent alignment across both training and testing phases. Then, we combine $\mathbf{x}_{\text{coarse}}$ with $\mathbf{x}_{\text{smooth}}$, using a certain probability $p_{\text{hybrid}}$ during training. Thus, the network is forced to be capable of generating textures even when adopting a weaker condition $\mathbf{x}_{\text{smooth}}$ in the fine stage during inference. Considering that the diffusion model first generates low-frequency information and then higher one~\cite{choi2022perception}, we also explore a condition-truncated sampling, as detailed discussed in Section~\ref{sec:ablation}, where we condition both maps for generation during the initial time steps and then exclusively utilize the smooth map for the remainder of the generation.

\input{my_figures/04_cluster.tex}
\input{my_figures/04_coarse.tex}

%% file: my_figures/04_cluster.tex
\begin{figure}[t]
\raggedright
    \includegraphics[trim={0cm 0cm 0cm 0cm},clip,width=\columnwidth]{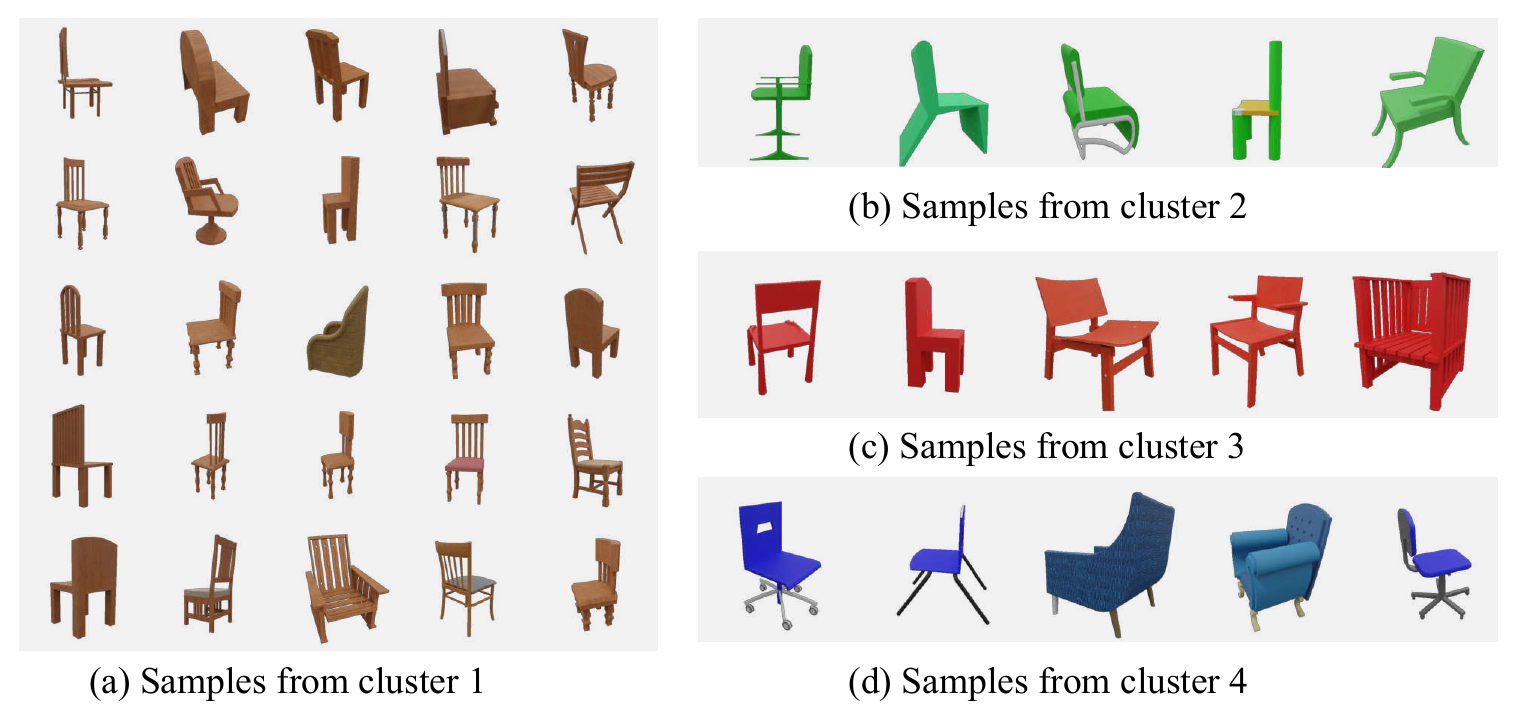}
    \caption{\textbf{Illustration of samples across various clusters.}
Shapes within a particular cluster exhibit analogous color styles.
However, there exists an imbalance in the quantity
of shapes among different clusters,
leading to challenges in unbiased synthesis.}
    \label{fig:cluster}
\end{figure}

%% file: my_figures/04_coarse.tex
\begin{figure}[t]
\raggedright
    \includegraphics[trim={0cm 0cm 0cm 0cm},clip,width=\columnwidth]{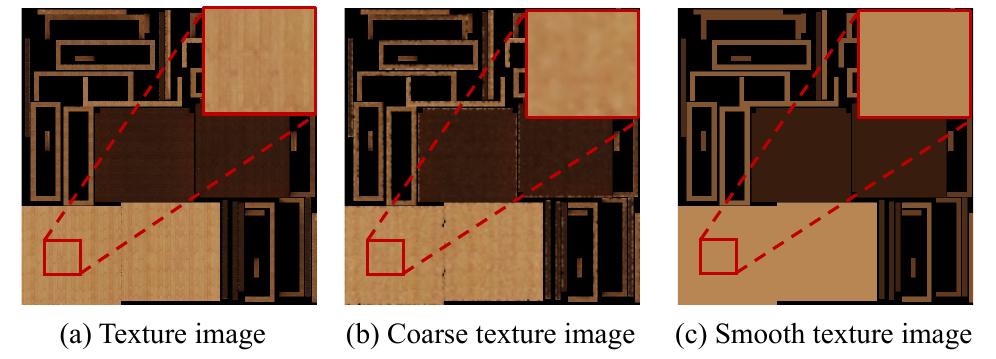}
\caption{\textbf{Varieties of texture images.}
(a) Original texture image enriched with high-frequency details,
(b) Coarse texture image,
and (c) Smooth texture image.}
    \label{fig:coarse}
\end{figure}

%% file: 05_experiments.tex
\section{Experiments} 
\label{sec:experiments}
\input{my_figures/04_result.tex}

\subsection{Datasets and Implementation Details}
We conduct experiments across four categories of the ShapeNet dataset 
\cite{shapenet2015}, 
namely, chair, table, car, and bench. 
Before training, we use an open-source UV-Atlas tool 
\cite{JonathanYoung18} 
to generate the UV map and pre-process the dataset 
to obtain the shape maps and ground-truth texture images. 
We sample $K=4096$ points in our coarse stage 
and synthesize a texture image in the UV space 
with a resolution of $512\times512$ for the fine stage. 
For training the diffusion models, akin to \cite{anciukevivcius2022renderdiffusion}, 
we predict the clean signals. 
This approach provides more stable training 
than predicting the noise component as recommended by \cite{ho2020denoising}. 
We employ the cosine noise scheduling \cite{nichol2021improved} 
ranging from $0.0001$ to $0.02$ 
over $1,024$ time steps for both stages. 
In the fine stage, we leverage the noise scaling strategy from \cite{chen2023importance} 
and also incorporate a rendering loss. 
Specifically, we randomly select four views 
and render the mesh using the predicted UV map 
and the ground-truth UV map to produce $1024\times1024$ images. 
Subsequently, we crop $224\times224$ patches from these images 
and compute the corresponding $L_1$ loss. 
For the hybrid condition in the fine stage, we use $p_{\text{hybrid}}=0.3$ and sweep over condition-truncated time $t_c$ (see Figure~\ref{fig:metric_drop}). 

\paragraph{Compared methods.} We compare our method with state-of-the-art approaches, including Texturify~\cite{siddiqui2022texturify}, Texture Fields~\cite{oechsle2019texture}, and PVD~\cite{zhou20213d}, in the context of unconditional texture generation. PVD was originally designed for point cloud generation instead of texture generation, so it cannot be directly compared with our approach. To facilitate comparison, we modify this framework to learn RGB values from the sampled point cloud. This extension, referred to as ``PVD-Tex'', serves as a baseline for directly learning texture in the point space using the diffusion model.  

\subsection{Unconditional Texture Generation} 

\paragraph{Gallary.} First, we showcase our texture generation results for each category without additional conditions. 
The results in Figure~\ref{fig:result} and Figure~\ref{fig:teasing} demonstrate the remarkable performance of our method, which is able to generate appealing textures with fine details and preserve the geometric structures.
Note that our method is compatible with meshes of diverse topology and intricate geometric details. 

\paragraph{Qualitative comparisons.}
As depicted in Figure~\ref{fig:compare}, our approach excels at generating high-quality textures while maintaining the geometric intricacies of the input mesh. Firstly, the results of Texture Fields~\cite{oechsle2019texture} appear deficient in high-frequency details. Additionally, while PVD-Tex~\cite{zhou20213d} is able to produce spatially varying colors to some extent (e.g., the car in Figure~\ref{fig:compare}), it falls short in synthesizing intricate high-frequency details. Lastly, even though Texturify~\cite{siddiqui2022texturify} 
demonstrates an ability to generate finer textures, 
it compromises on preserving slender structures 
(\ie chair).

\paragraph{Quantitative comparisons.}
To quantitatively compare with existing works, we follow~\cite{siddiqui2022texturify} and assess the generation quality using Frechet Inception Distance (FID)\cite{heusel2017gans} and Kernel Inception Distance (KID)\cite{binkowski2018demystifying}, metrics that are widely used for evaluating image generation models. To this end, we render 512 $\times$ 512 images from each generated textured mesh and ground-truth textured mesh using four distinct camera views.
Table~\ref{table:comparison} presents the quantitative comparisons with current approaches, revealing that our method surpasses existing works.

\input{my_figures/04_compare.tex}

\subsection{Conditional Texture Generation} 
In addition, we demonstrate the capability of our framework to synthesize textures conditioned on either text prompts or a single-view image. We conduct our experiments on the chair and table categories. For the text condition, we utilize text-shape pairs as provided in~\cite{chen2019text2shape} (with additional corrections for text accuracy). For the image condition, we randomly render a view from the ground-truth mesh. To infuse the network with condition-specific information, we use the pre-trained vision-language model CLIP~\cite{radford2021learning} to extract the corresponding embedding from either the image or the text. This embedding is then fed into a simple MLP to incorporate the information into the diffusion model as a condition for both training and inference. As depicted in Figure~\ref{fig:text}, our method succeeds in generating textures that align well with the given text descriptions or images.
We also compare our approach with Text2Mesh~\cite{michel2022text2mesh}, a test-time optimization method. Text2Mesh takes around 10 minutes per instance, while ours only requires 30 seconds. Importantly, to adapt to our task which requires preserving the mesh geometry, we freeze the geometry deformation branch of Text2Mesh to generate only colors. As shown in Figure~\ref{fig:text2mesh}, Text2Mesh is bound by vertex resolution, while our method can deliver detailed visuals even on low-resolution meshes, offering a distinct advantage in graphics.

\input{my_figures/05_text.tex}
\input{my_figures/text2mesh.tex}

\begin{table*}[t]
    \centering
    \begin{tabular}{l|ll|ll|ll|ll}
    \hline
        & \multicolumn{2}{c|}{Chair} & \multicolumn{2}{c|}{Car} & \multicolumn{2}{c|}{Table} & \multicolumn{2}{c}{Bench}\\ 
        \hline
       Methods & FID $\downarrow$     & KID  $\downarrow$   & FID $\downarrow$     & KID   $\downarrow$  & FID $\downarrow$     & KID   $\downarrow$ & FID $\downarrow$     & KID  $\downarrow$       \\ \hline
    Texture Fields~\cite{oechsle2019texture} & 24.24  & 1.07     & 156.38  & 13.64 & 68.96  & 4.20 & 62.71  & 2.96\\ 
    Texturify~\cite{siddiqui2022texturify}      & 27.80  & 1.32     & 73.16  & 4.71  & -  & - & -  & - \\
    PVD-Tex~\cite{zhou20213d}      & 15.52  & 0.62        & 59.47   & 3.74 & 16.12  & 0.55 & 28.94  & 0.39\\
    Ours      & \bf{9.88}  & \bf{0.22} & \bf{26.89}  & \bf{0.68}  & \bf{9.63}  & \bf{0.15} & \bf{23.09}  & \bf{0.15} \\
    \hline
    \end{tabular}
    \caption{\textbf{Quantitative comparisons with existing works.} Ours outperforms other approaches on both FID~\cite{heusel2017gans} and KID ($\times10^{2}$)~\cite{binkowski2018demystifying} .}
\label{table:comparison}
\end{table*}

\input{my_figures/05_ablation.tex}

\subsection{Ablation Studies} 
\label{sec:ablation}

\paragraph{Coarse-to-fine diffusion.}
To manifest the effectiveness of our coarse-to-fine diffusion strategy, we conduct experiments on two baselines ``w/o coarse stage'' and ``w/o fine stage''. The ``w/o coarse stage'' configuration indicates that we directly generate the UV texture image using the fine stage, bypassing the initialization from the coarse stage. 
The ``w/o fine stage'' configuration signifies the result of the coarse stage, after assigning colors to the uncolored pixels using KNN interpolation. 
In both scenarios, the model produces inferior outcomes relative to our full model, as shown in Table~\ref{table:ablation} and Figure~\ref{fig:ablation}. Absent the coarse stage, the generated result suffers from noticeably inconsistent colors. Without the fine stage, the results are over-smoothed.

\paragraph{Hybrid condition.}
As shown in Table~\ref{table:ablation} and Figure~\ref{fig:ablation} ``coarse condition'', the fine stage cannot generate high-quality textures when it is exclusively conditioned on the coarse map. Further, the generated quality remains unsatisfactory if we apply a hybrid condition entirely during training (\ie, $p_{\text{hybrid}}=1.0$), as shown in ``$p_{\text{hybrid}}=1.0$''. We attribute this to the network's propensity to depend solely on the coarse map during training and generation. In contrast, our full model (\ie, hybrid condition with $p_{\text{hybrid}}=0.3$) achieves significant improvement both qualitatively and quantitatively, as presented in Table~\ref{table:ablation} and Figure~\ref{fig:ablation} ``ours''. Furthermore, we also sweep over the effect of condition-truncated time $t_c$ for inference. As Figure~\ref{fig:metric_drop} shows, $t_c=0.4$ strikes a sweet point. 
The model utilizes information from both conditions maps to better generate low-frequency components during the first $40\%$ of the sampling timesteps. After that, the coarse map $\mathbf{x}_{\text{coarse}}$ is dropped, and the model further focuses on fine-grained detail generation without relying on the coarse map. 

\begin{table}[t]
    \centering

    \begin{tabular}{ccc}
    \hline
       & FID $\downarrow$     & KID $\downarrow$   \\ 
    \hline
    w/o fine stage & 17.88  & 0.76    \\
    w/o coarse stage      & 15.11  & 0.49     \\
    coarse condition  & 14.93  & 0.56     \\ 
    $p_{\text{hybrid}}=1.0$ & 15.25  & 0.59     \\
    Ours & \bf{9.88}  & \bf{0.22}     \\
    \hline
    \end{tabular}
    \vspace{2pt}
    \caption{\textbf{Ablation studies.} This table shows the effectiveness of each component in our proposed method.}
\label{table:ablation}
\end{table}

\paragraph{Style guidance.}
The style guidance is aimed at addressing the issue of insufficient diversity due to color distribution bias in the dataset. As a result, we do not utilize FID or KID for evaluation since they evaluate the distribution similarity between generated results and ground truth. However, the ground-truth distribution shows a strong bias towards particular colors, making these metrics inappropriate for assessing generative texture diversity. In contrast, we adopt LPIPS~\cite{zhang2018unreasonable} to measure the pairwise similarity among five textures generated by our method given the same input mesh, where a larger diversity in textures will lead to a higher LPIPS value. In Figure~\ref{fig:lpips}, we report the quantitative results for 500 shapes. Our approach achieves a higher LPIPS in most of the evaluated cases, indicating better diversity. Besides, as shown in Figure~\ref{fig:style}, we present the results of generating three textures randomly for the shapes without and with style guidance. In the latter case, we uniformly sample three style labels as style inputs. It is evident that without style guidance, the generated textures are nearly identical, and the color styles of different shapes tend to be similar. With the introduction of style guidance, however, the diversity of colors has significantly increased. We also conduct a comparison with other methods in terms of diversity, as well as a quality assessment through user studies, as shown in Table~\ref{table:user}. 

\begin{table}
\centering
\begin{tabular}{ccc}
\hline
Methods & Preference$\uparrow$  & LPIPS$\uparrow$  \\
\hline
Ours  & \bf{49.8\%} & 0.083 \\
Texturify~\cite{siddiqui2022texturify} & 15.9\% & \bf{0.086} \\
PVD-Tex~\cite{zhou20213d} & 29.2\% & 0.029 \\
Texture Fields~\cite{oechsle2019texture} & 5.1\% & 0.005 \\
\hline
\end{tabular}
\vspace{2pt}
\caption{\textbf{LPIPS and user study.} This table shows the average LPIPS and preference via user study in the chair category.}
\label{table:user}
\end{table}

\input{my_figures/08_drop.tex}
\input{my_figures/08_lpips.tex}
\input{my_figures/08_style.tex}

%% file: my_figures/04_result.tex
\begin{figure*}
\centering
    \includegraphics[width=\textwidth]{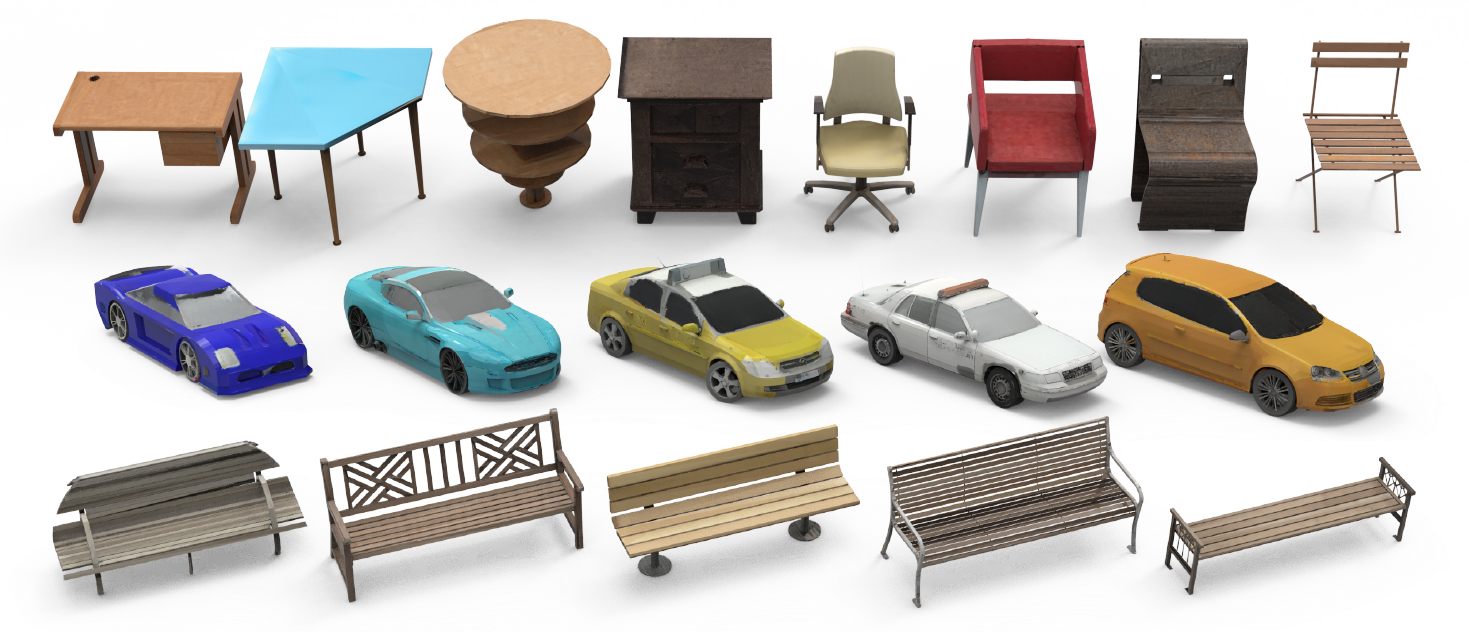}
    \caption{\textbf{Our texture synthesis results in different categories.} The results on chairs, tables, cars, and benches demonstrate that our approach can generate natural, vivid, and diverse textures, even though the given 3D shapes are challenging, such as cutouts on the bench.}
    \label{fig:result}
    % \vspace{0.4cm}
\end{figure*}

%% file: my_figures/04_compare.tex
\begin{figure}
    \includegraphics[trim={0cm 0cm 0cm 0cm},clip,width=0.99\columnwidth]{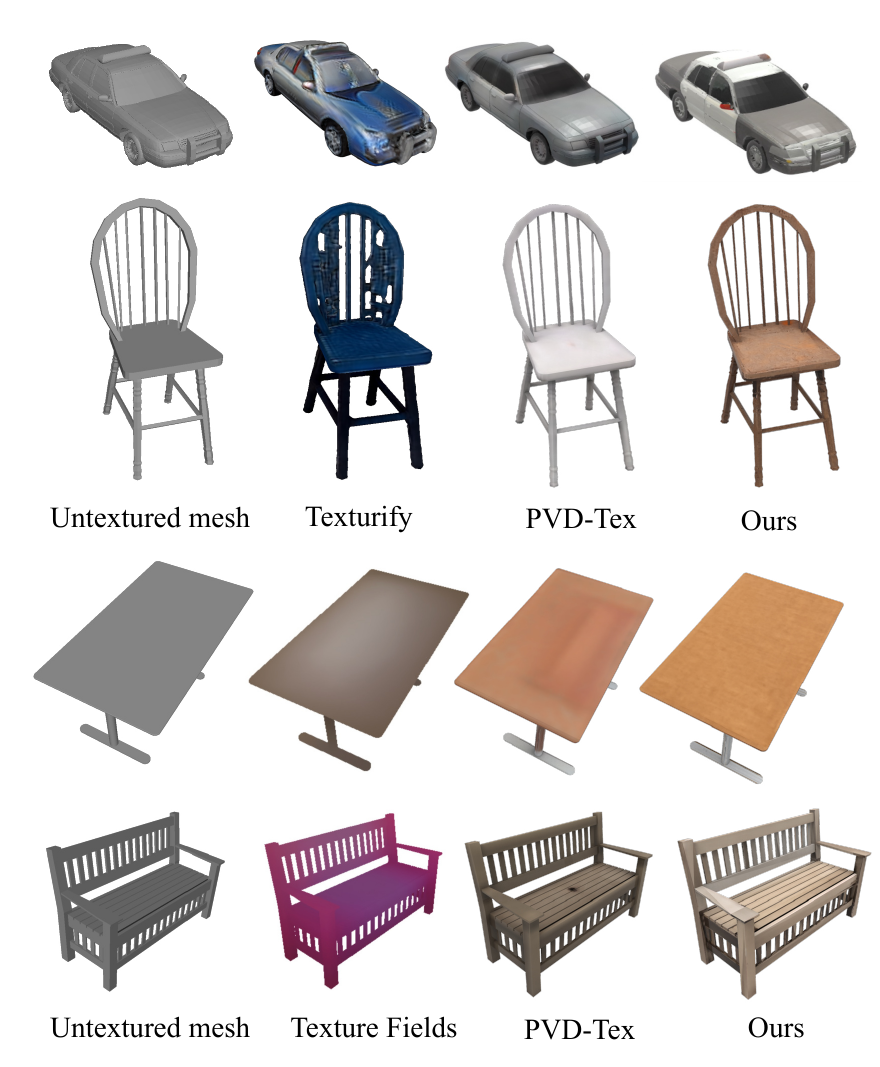}
    \caption{\textbf{Qualitative comparisons with existing works.} Given pure 3D shapes, Texturify~\cite{siddiqui2022texturify} generates textures on the surface of 3D shape but tends to damage the topology; Texture fields~\cite{oechsle2019texture} and PVD-Tex~\cite{zhou20213d} produce textures with limited details. On the contrary, our approach faithfully preserves topology and produces realistic appearances.}
    \label{fig:compare}
\end{figure}

%% file: my_figures/05_text.tex
\begin{figure}
    \includegraphics[trim={0cm 0cm 0cm 0cm},clip,width=0.99\columnwidth]{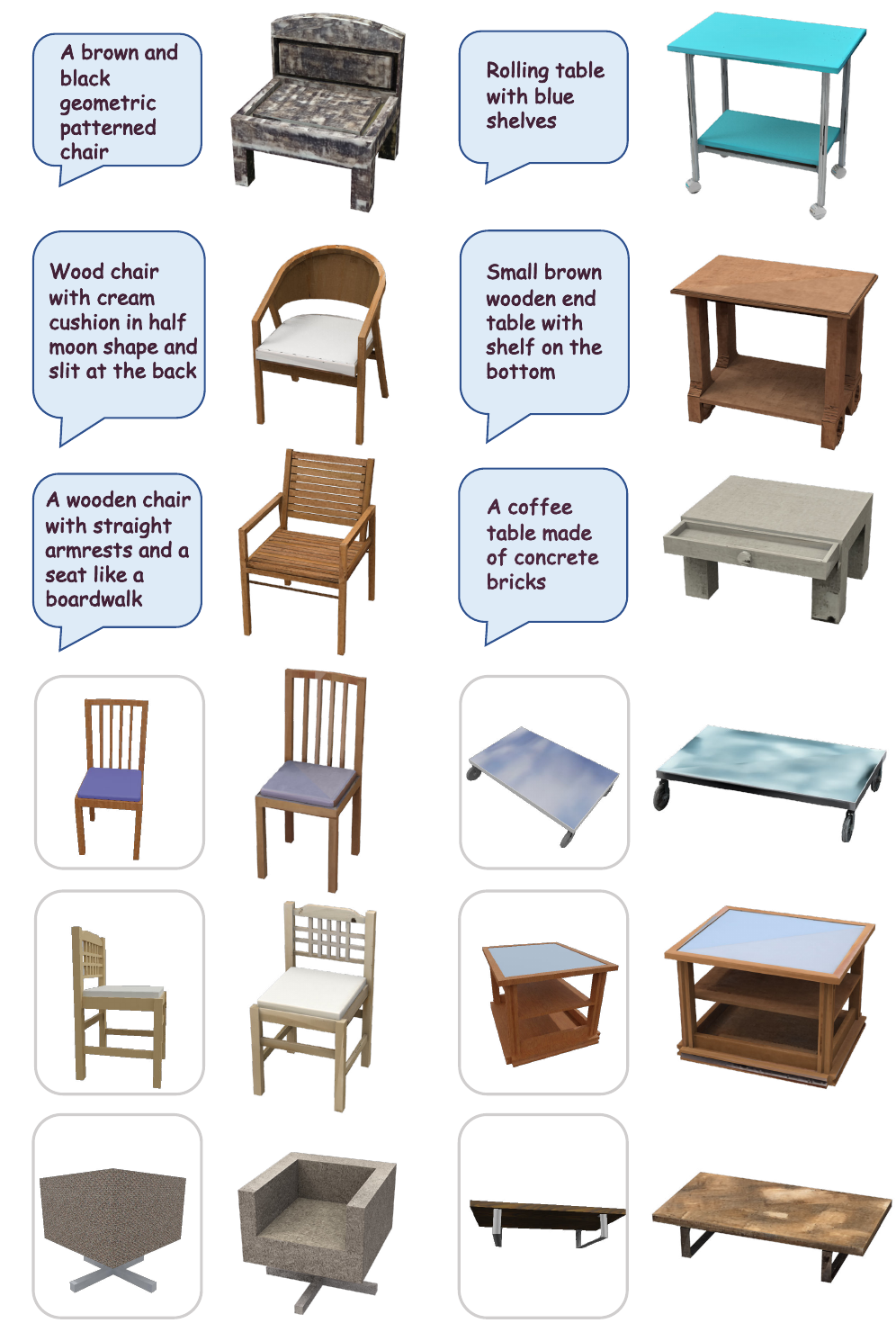}
    \caption{\textbf{Results of conditional texture generation.} Our method is adaptable to craft textures guided by text descriptions 
(rows 1-3) or single-view images (rows 4-6).}
    \label{fig:text}
\end{figure}

%% file: my_figures/text2mesh.tex
\begin{figure}
\centering
    \includegraphics[trim={0cm 0cm 0cm 0cm},clip,width=0.9\columnwidth]{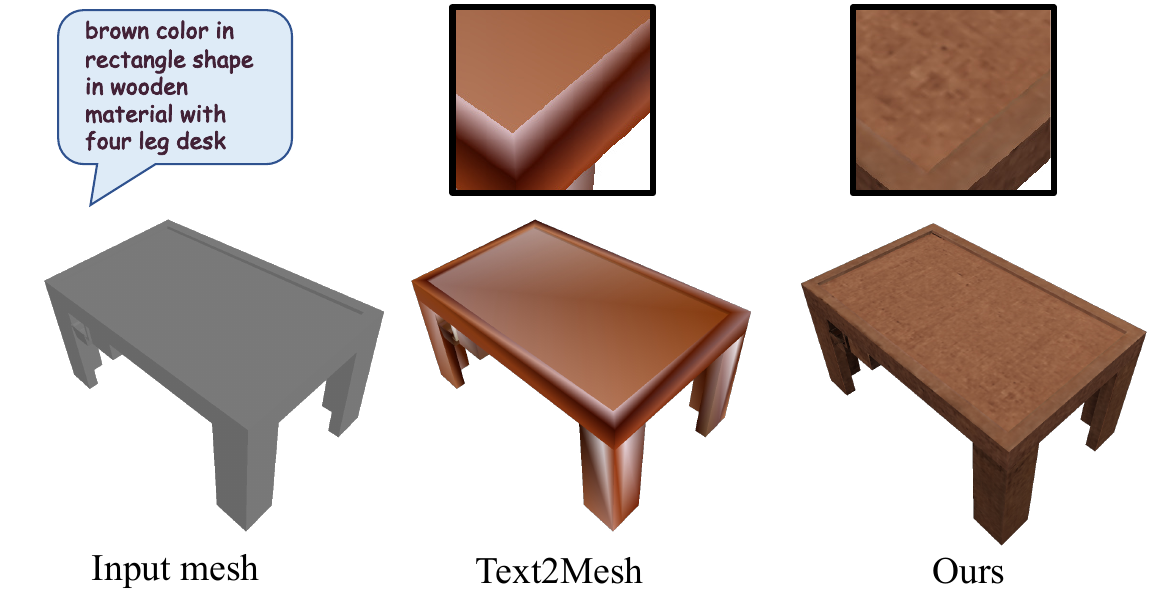}
    \caption{\textbf{Comparison with Text2Mesh.} We freeze the geometry deformation branch of Text2Mesh to adapt to our task. Ours can generate high-frequency details for low-resolution mesh.}
    \label{fig:text2mesh}
\end{figure}

%% file: my_figures/05_ablation.tex
\begin{figure}
\centering
    \includegraphics[trim={0cm 0cm 0cm 0cm},clip,width=0.85\columnwidth]{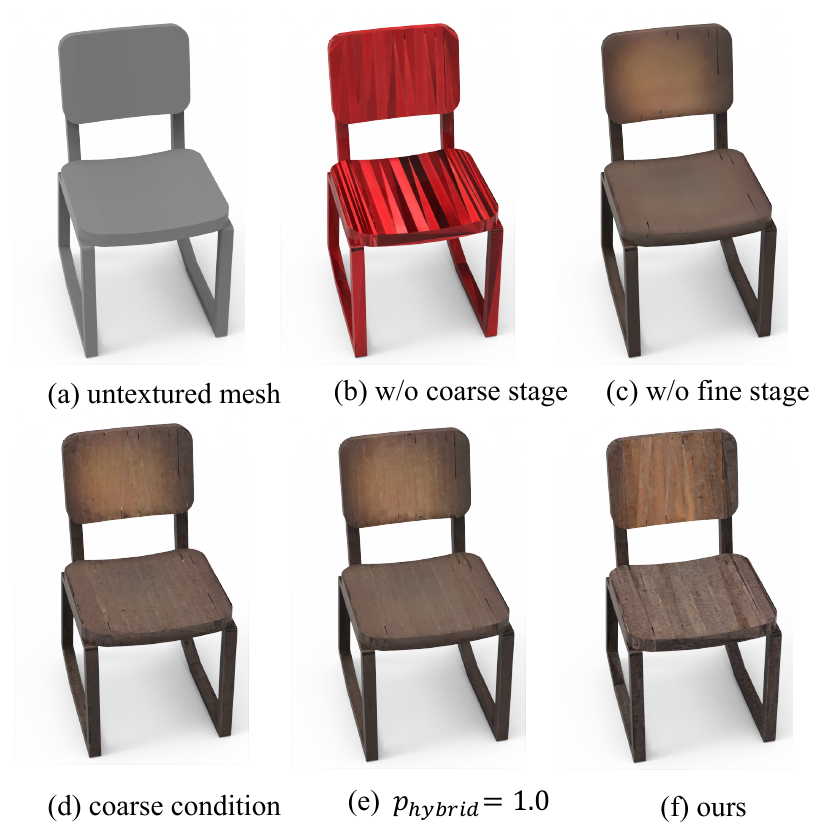}
    \caption{\textbf{Ablation studies.} Without our two-stage generation pipeline or hybrid condition designs, the generated results are inconsistent with the 3D shape or lack high-frequency details.}
    \label{fig:ablation}
\end{figure}

%% file: my_figures/08_drop.tex
\begin{figure}[t]
    % \vspace{-10pt}
    \centering
    \includegraphics[trim={0cm 0cm 0cm 0cm},clip,width=0.9\columnwidth]{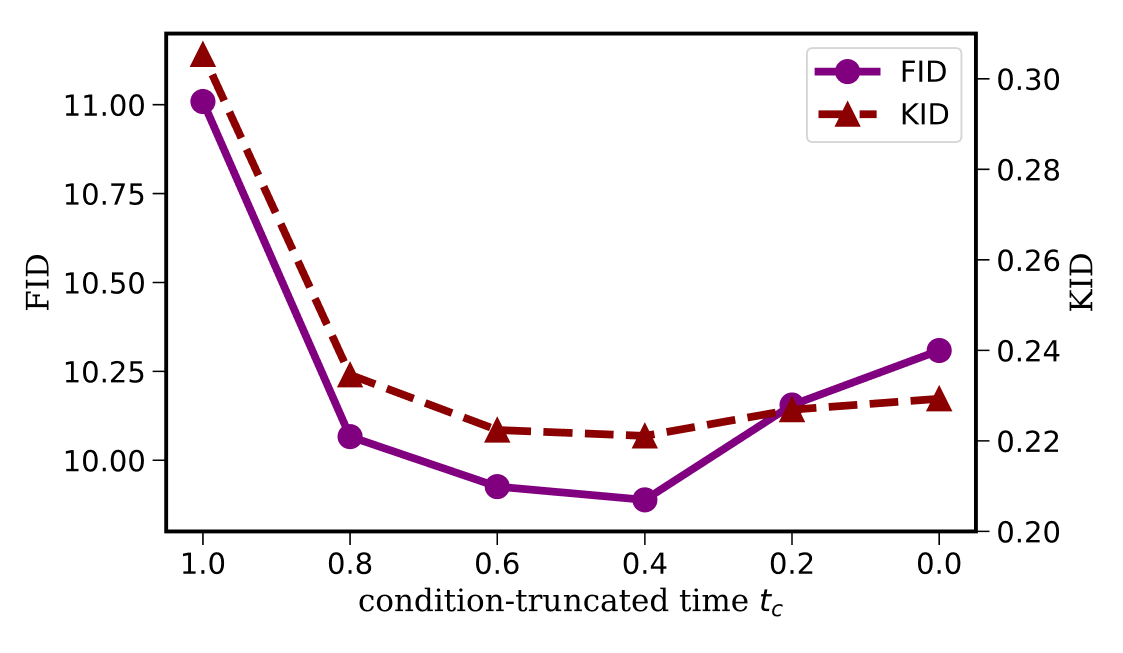}
\caption{\textbf{Examination of condition-truncated time.} 
The horizontal axis denotes the percentage of time points 
when the coarse texture image is conditioned during inference. 
The left vertical axis indicates the FID value, 
while the right vertical axis shows the KID value ($\times 10^{2}$).}

    \label{fig:metric_drop}
    \vspace{-0.2cm}
\end{figure}

%% file: my_figures/08_lpips.tex
\begin{figure}
\centering
    \includegraphics[trim={0cm 0cm 0cm 0cm},clip,width=0.9\columnwidth]{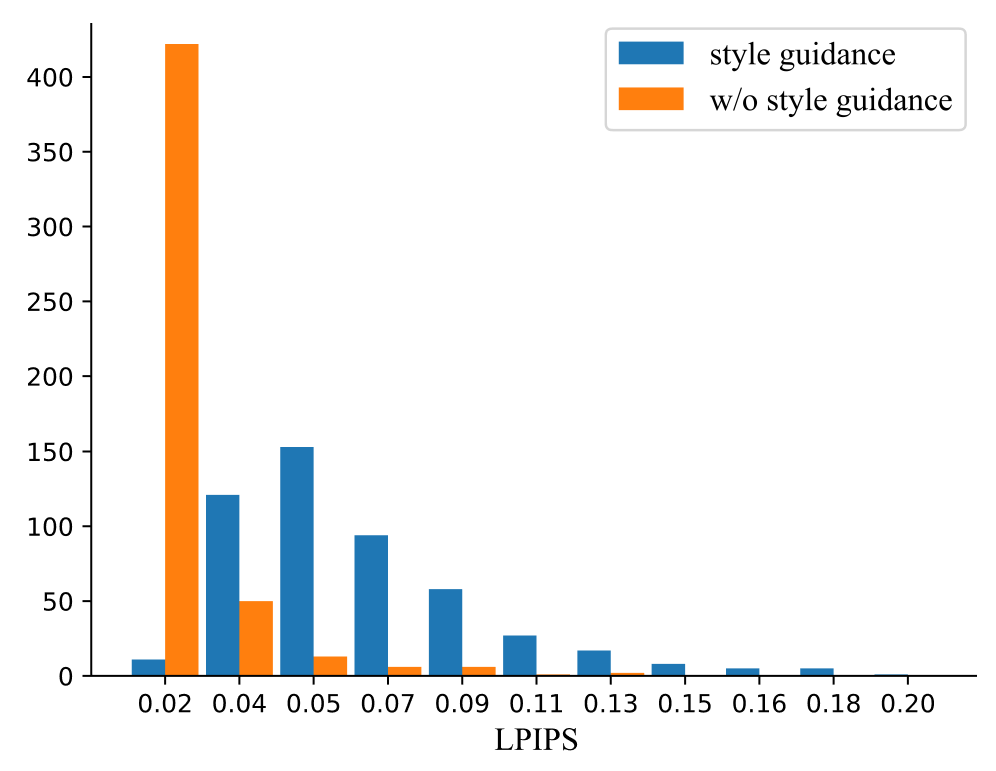}
    \caption{\textbf{Diversity measurement.} Utilizing style guidance leads to larger LPIPS scores, indicating enhanced generation diversity.}

    \label{fig:lpips}
\end{figure}

%% file: my_figures/08_style.tex
\begin{figure}
\centering
    \includegraphics[trim={0cm 0cm 0cm 0cm},clip,width=0.9\columnwidth]{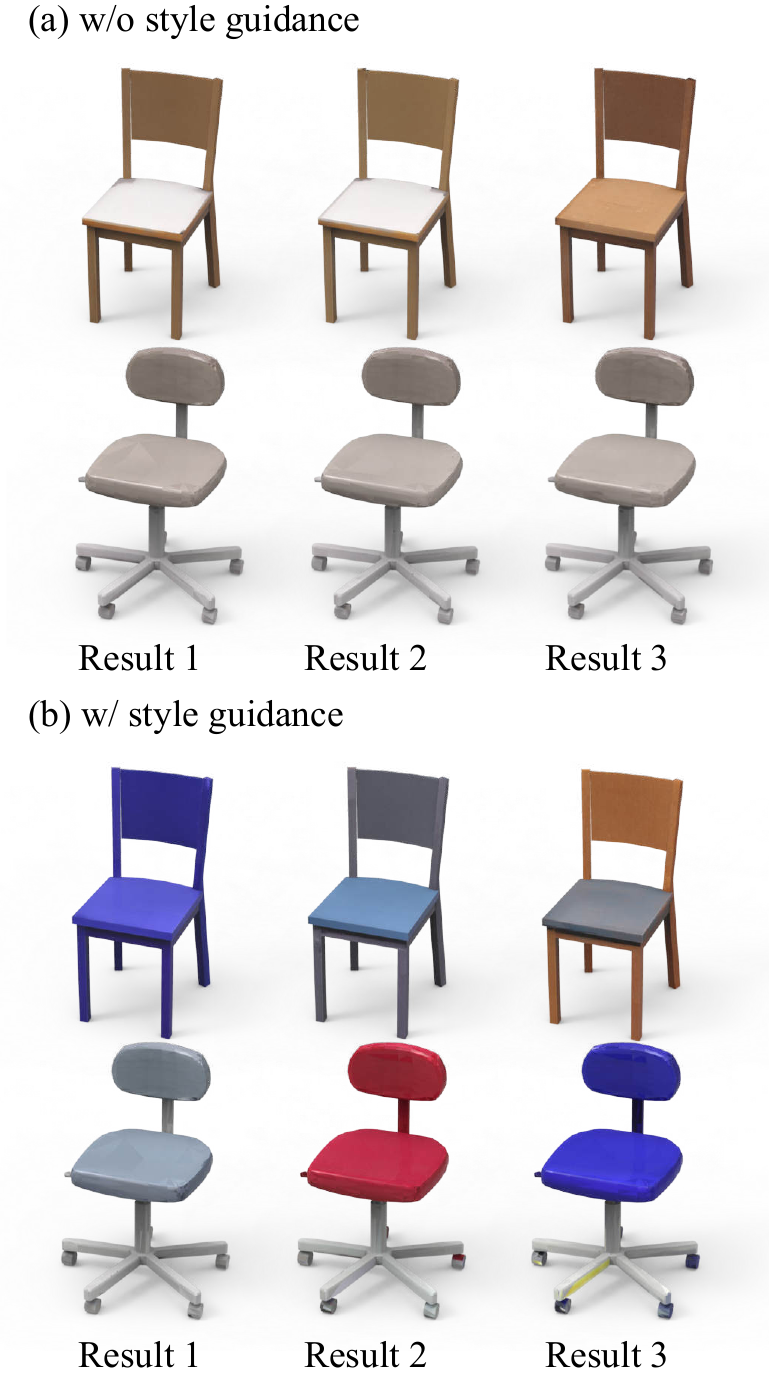}
    \caption{\textbf{Efficacy of style guidance.} The model without style guidance produces consistent colors across different random seeds (rows 1-2). In contrast, with style guidance, diverse outcomes emerge (rows 3-4).}
    \label{fig:style}
\end{figure}

%% file: 06_conclusion.tex
\begin{figure}
\centering
    \includegraphics[trim={0cm 0cm 0cm 0cm},clip,width=0.9\columnwidth]{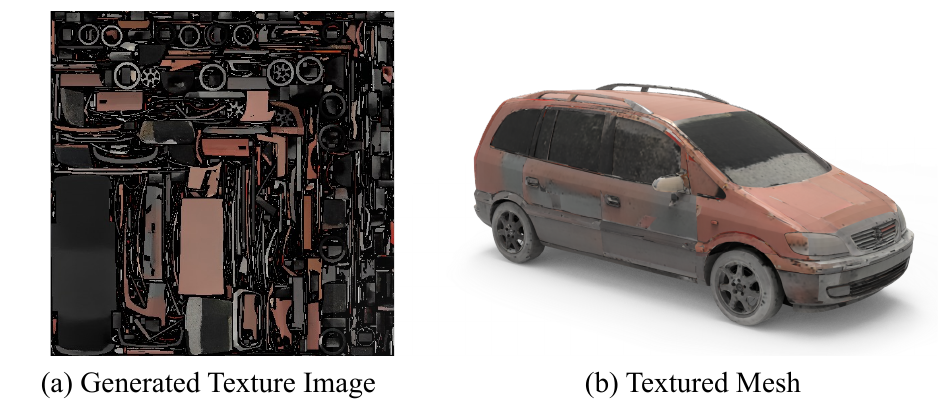}
    \caption{\textbf{Failure case.} Our approach still struggles to generate seamless results when there are too many fragmented cuts of the UV map.}
    \label{fig:fail}
    \vspace{-0.4cm}
\end{figure}

\section{Discussion, Limitations, and Conclusions}

This paper presents Point-UV diffusion, a brand-new framework that employs a coarse-to-fine pipeline to generate textures for 3D meshes. We begin with a 3D diffusion model to synthesize low-frequency texture components from point clouds, which maintain 3D consistency. We then refine the textures using a 2D UV-space diffusion model. Our method is compatible with meshes with arbitrary topology and can
faithfully preserve the geometry structure. We further demonstrate the flexibility of our framework by extending it to conditional generative models.

Despite its merits, our method has inherent limitations. 
Similar to other methodologies relying on 3D data for training, 
our technique is upper-bounded by the scope and diversity 
of current 3D datasets. 
This restriction poses challenges in generating textures 
that parallel the depth of effects seen in 2D image synthesis.  Moreover, our method's efficacy relies on  
the quality of UV mapping. 
Our approach faces difficulties in rendering high-quality results for meshes where the UV mapping produces excessive fragmented cuts, 
resulting in fragmented artifacts. This phenomenon is commonly observed in the car category, as shown in the Figure~\ref{fig:fail}. 
We believe the emergence of larger and more 
diverse 3D datasets would be helpful for  generating superior-quality textures. 
Further advancements in UV parameterization would also be beneficial 
in augmenting our method's consistency. 

\section{Acknowledgements}
This work has been supported by Hong Kong Research Grant Council - Early Career Scheme (Grant No. 27209621), General Research Fund Scheme (Grant No. 17202422), and RGC matching fund scheme (RMGS). Part of the described research work is conducted in the JC STEM Lab of Robotics for Soft Materials funded by The Hong Kong Jockey Club Charities Trust.

%% file: supplement.tex
%%%%%%%%% TITLE
% \title{Appendix: Texture Generation on 3D Meshes with Point-UV Diffusion}

% \maketitle

\input{./supp_sections/ddpm.tex}

\input{./supp_sections/method.tex}
\input{./supp_sections/implementations.tex}

% \clearpage
% {\small
% \bibliographystyle{ieee_fullname}
% \bibliography{egbib}
% }

% \end{document}

%% file: supp_sections/ddpm.tex
\section{Denoising Diffusion Model}
\label{sec:diffusion}

\paragraph{Forward process.}
Denoising diffusion probabilistic model (DDPM)~\cite{nichol2021improved} learns the data distribution by introducing a series of latent variables and matching the joint distribution.
The model starts with a sample from the data distribution, denoted as $x_0 \sim q(\mathbf{x}_0)$, and a forward process $q\left(\mathbf{x}_{1: T} \mid \mathbf{x}_0\right)=\prod_{t=1}^T q\left(\mathbf{x}_t \mid \mathbf{x}_{t-1}\right)$ progressively perturbs the data with Gaussian kernels $q\left(\mathbf{x}_t \mid \mathbf{x}_{t-1}\right):=\mathcal{N}\left(\sqrt{1-\beta_t} \mathbf{x}_{t-1}, \beta_t \mathbf{I}\right)$, producing increasingly noisy latent variables $\mathbf{x}_{1}, \mathbf{x}_{2}, ..., \mathbf{x}_{T}$. Notably, $x_t$ can be directly sampled from $x_0$ thanks to the closed form:
\begin{equation}
q\left(\mathbf{x}_t \mid \mathbf{x}_0\right)=\mathcal{N}\left(\mathbf{x}_t ; \sqrt{\bar{\alpha}_t} \mathbf{x}_0,\left(1-\bar{\alpha}_t\right) \mathbf{I}\right),
\end{equation}
where $\alpha_t:=1-\beta_t$ and $\bar{\alpha}_t:=\prod_{s=1}^t \alpha_s$.
In general, the forward process variances $\beta_t$ are fixed and increased linearly from $\beta_1=10^{-4}$ to $\beta_T=0.02$. Besides, $T$ should be large ({\eg}, 1000) enough to ensure $q\left(\mathbf{x}_T \mid \mathbf{x}_0\right)\approx \mathcal{N}(0, \mathbf{I})$. 
The diffusion model aims to model the joint distribution $q\left(\mathbf{x}_{0:T}\right)$, which includes a tractable sampling path for the marginal distribution $q\left(\mathbf{x}_{0}\right)$.

\paragraph{Reverse process and optimization.}
To learn how to reverse the forward process, the diffusion model defines a parameterized Markov chain with parameterized transition kernels:
\begin{equation}
\begin{aligned}
    p_\theta\left(\mathbf{x}_{0: T}\right)
    &:=p\left(\mathbf{x}_T\right) \prod_{t=1}^T p_\theta\left(\mathbf{x}_{t-1} \mid \mathbf{x}_t\right),\\
    p_\theta\left(\mathbf{x}_{t-1} \mid \mathbf{x}_t\right)
    &:=\mathcal{N}\left(\mathbf{x}_{t-1} ; \boldsymbol{\mu}_\theta\left(\mathbf{x}_t, t\right), \boldsymbol{\Sigma}_\theta\left(\mathbf{x}_t, t\right)\right),
\end{aligned}
\end{equation}
where $\boldsymbol{\mu}_\theta$ and $\boldsymbol{\Sigma}_\theta$ are optimized. The training is performed by optimizing a variational bound of negative log likelihood: 
{\small
\begin{equation}
\begin{aligned}
   \mathrm{E}_{q\left(\mathbf{x}_0\right)}\left[-\log p_\theta\left(\mathbf{x}_0\right)\right] 
   \leq \mathrm{E}_{q\left(\mathbf{x}_{0: T}\right)}\left[-\log \frac{p_\theta\left(\mathbf{x}_{0: T}\right)}{q\left(\mathbf{x}_{1: T} \mid \mathbf{x}_0\right)}\right] =: L.
\end{aligned}
\end{equation}
}

The loss term $L$ can be rewritten as:
\\
% \begin{adjustbox}{max width=0.47\textwidth}
{\small
\begin{equation}
\begin{aligned}
&L=\mathbb{E}_q[\underbrace{D_{\mathrm{KL}}\left(q\left(\mathbf{x}_T \mid \mathbf{x}_0\right) \| p\left(\mathbf{x}_T\right)\right)}_{L_T}\\
&+\sum_{t>1} \underbrace{D_{\mathrm{KL}}\left(q\left(\mathbf{x}_{t-1} \mid \mathbf{x}_t, \mathbf{x}_0\right) \| p_\theta\left(\mathbf{x}_{t-1} \mid \mathbf{x}_t\right)\right)}_{L_{t-1}} \underbrace{-\log p_\theta\left(\mathbf{x}_0 \mid \mathbf{x}_1\right)}_{L_0}].
\end{aligned}
\end{equation}
}

In practice, the core optimization terms are $L_{t-1} (t>1)$ that can be analytically calculated since both two terms compared in the KL divergence are Gaussians, {\ie},:
\begin{equation}
\begin{aligned}
&q\left(\mathbf{x}_{t-1} \mid \mathbf{x}_t, \mathbf{x}_0\right)=\mathcal{N}\left(\mathbf{x}_{t-1} ; \tilde{\boldsymbol{\mu}}_t\left(\mathbf{x}_t, \mathbf{x}_0\right), \tilde{\beta}_t \mathbf{I}\right), \\
&p_\theta\left(\mathbf{x}_{t-1} \mid \mathbf{x}_t\right):=\mathcal{N}\left(\mathbf{x}_{t-1} ; \boldsymbol{\mu}_\theta\left(\mathbf{x}_t, t\right), \boldsymbol{\Sigma}_\theta\left(\mathbf{x}_t, t\right)\right),
\end{aligned}
\end{equation}
where $\tilde{\boldsymbol{\mu}}_t\left(\mathbf{x}_t, \mathbf{x}_0\right):=\frac{\sqrt{\bar{\alpha}_{t-1}} \beta_t}{1-\bar{\alpha}_t} \mathbf{x}_0+\frac{\sqrt{\alpha_t}\left(1-\bar{\alpha}_{t-1}\right)}{1-\bar{\alpha}_t} \mathbf{x}_t \quad$ and $\quad \tilde{\beta}_t:=\frac{1-\bar{\alpha}_{t-1}}{1-\bar{\alpha}_t} \beta_t$.
DDPM~\cite{ho2020denoising} fix $\boldsymbol{\Sigma}_\theta\left(\mathbf{x}_t, t\right)=\sigma_t^2 \mathbf{I}$ during training, where $\sigma_t^2$ is set to be $\beta_t$ or $\tilde{\beta}_t$. In our experiments, we set $\sigma_t^2=\beta_t$.

\paragraph{Loss function.}
Models can be trained in the reverse process to predict the mean value of $\mathbf{x}_{t-1}$, \ie $\boldsymbol{\mu}_t$. Alternatively, by modifying the parameterization, we can also train it to predict $\mathbf{x}_0$ or $\boldsymbol{\epsilon}$, as illustrated in~\cite{ho2020denoising}. The original DDPM~\cite{ho2020denoising} predicts $\boldsymbol{\epsilon}$. In this case, through reparameterization trick \cite{kingma2013auto} and empirical simplification \cite{ho2020denoising}, the final training term is performed as follows:
\begin{equation}
L_{\text {simple }}(\theta):=\mathbb{E}_{t, \mathbf{x}_0, \epsilon}\left[\left\|\boldsymbol{\epsilon}-\boldsymbol{\epsilon}_\theta\left(\sqrt{\bar{\alpha}_t} \mathbf{x}_0+\sqrt{1-\bar{\alpha}_t} \boldsymbol{\epsilon}, t\right)\right\|^2_2\right],
\end{equation}
where $\epsilon\sim \mathcal{N}(0, \mathbf{I})$ and $t$ is uniformly sampled between $1$ and $T$.
However, in our experiments, we predict $\mathbf{x}_0$ since we find it results in more stable training. The loss function is modified as:

\begin{equation}
\label{eq:7}
L_{\text{x}_0}(\theta):=\mathbb{E}_{t, \mathbf{x}_0, \epsilon}\left[\left\|\mathbf{x}_0-G_\theta\left(\sqrt{\bar{\alpha}_t} \mathbf{x}_0+\sqrt{1-\bar{\alpha}_t} \boldsymbol{\epsilon}, t\right)\right\|^2_2\right],
\end{equation}
where $G_\theta$ is the network to be optimized.

\paragraph{Sampling.}
After training, started from an initial noise map $x_T\sim p(\mathbf{x}_T)=\mathcal{N}(0, \mathbf{I})$, new images can be then generated via iteratively sampling from $p_\theta\left(\mathbf{x}_{t-1} \mid \mathbf{x}_t\right)$, using the following equation:
\begin{equation}
\begin{aligned}
\mathbf{x}_{t-1}=\tilde{\boldsymbol{\mu}}_t+\sigma_t \mathbf{z}, 
\end{aligned}
\end{equation}
where $\mathbf{z} \sim \mathcal{N}(\mathbf{0}, \mathbf{I})$. 
In particular, we predict the $x_0$, so that the sampling process can be modified as:
\begin{equation}
\begin{aligned}
\mathbf{x}_{t-1}=
&\frac{\sqrt{\bar{\alpha}_{t-1}} \beta_t}{1-\bar{\alpha}_t} \mathbf{\hat{x}}_0+\frac{\sqrt{\alpha_t}\left(1-\bar{\alpha}_{t-1}\right)}{1-\bar{\alpha}_t} \mathbf{x}_t
+\sigma_t \mathbf{z} \\
=&\frac{\sqrt{\bar{\alpha}_{t-1}} \beta_t}{1-\bar{\alpha}_t} G_{\theta}\left(\mathbf{x}_t, t\right)+\frac{\sqrt{\alpha_t}\left(1-\bar{\alpha}_{t-1}\right)}{1-\bar{\alpha}_t} \mathbf{x}_t
+\sigma_t \mathbf{z}.
\end{aligned}
\end{equation}

%% file: supp_sections/method.tex
\section{Method}
\label{sec:sup_method}
\subsection{Basic Point-UV Diffusion}
In this section, we introduce the basic training process of our model, which \textit{\textbf{does not} include style guidance and hybrid conditioning}.
Our nets are denoted by $G^1_{\theta_1}$ and $G^2_{\theta_2}$ for the coarse stage and fine stage. As we have mentioned in Section \ref{sec:diffusion}, we train the models to predict clean signals instead of noise components. 

We summarize our training procedure in Algorithm~\ref{alg1}. In this algorithm flowchart, $q$ represents the true texture distribution that we aim to estimate, while $\mathbf{z}_0$ and $\mathbf{x}_0$ represent the ground-truth point cloud color and texture image, respectively. $L_{\text{coarse}}$ and $L_{\text{fine}}$ represent the loss functions for the two stages, which will be described in detail below. It should be noted that the training of the fine stage can be parallel to the coarse stage, as it does not depend on the training of the coarse stage. 

\paragraph{Loss functions.}
For the coarse stage, we simply adopt the loss function we described in Equation~\ref{eq:7}, \ie:
\begin{equation}
\begin{aligned}
L_{\text{coarse}}:=\mathbb{E}_{t, \mathbf{z}_0, \epsilon}\left[\left\|\mathbf{z}_0-G^1_{\theta_1}\left(\mathbf{z}_t, t, \mathbf{x}_{\text{shape}}, \mathbf{z}_{\text{coord}}\right)\right\|^2_2\right],
\end{aligned}
\end{equation}
where $G^1_{\theta_1}$ is the coarse network and $\mathbf{z}_t = \sqrt{\bar{\alpha}_t} \mathbf{z}_0+\sqrt{1-\bar{\alpha}_t} \boldsymbol{\epsilon}$ is the noisy point cloud color.

For the fine stage, in addition to using the basic diffusion loss, \ie:
\begin{equation}
L_{\text{basic}}:=\mathbb{E}_{t, \mathbf{x}_0, \epsilon}\left[\left\|\mathbf{x}_0-G^2_{\theta_2}\left(\mathbf{x}_t, t, \mathbf{x}_{\text{shape}}, \mathbf{x}_{\text{coarse}}\right)\right\|^2_2\right],
\end{equation}
where $G^2_{\theta_2}$ is the fine network and $\mathbf{x}_t = \sqrt{\bar{\alpha}_t} \mathbf{x}_0+\sqrt{1-\bar{\alpha}_t} \boldsymbol{\epsilon}$ is the noisy texture image.
we also utilize differentiable rendering to render the generated textured mesh and the ground-truth mesh, then calculate the $L_1$ rendering loss between rendered images. Specifically, during training, by using Nvdiffrast~\cite{laine2020modular}, we randomly select four views and render the mesh based on the predicted UV map and ground-truth UV map to obtain four 1024$\times$1024 images. We then crop patches of size 224$\times$224 from each of these images to obtain $\hat{\mathbf{x}}_{\text{render}}\in\mathbb{R}^{4\times3\times224\times224}$ and $\mathbf{x}_{\text{render}}\in\mathbb{R}^{4\times3\times224\times224}$. The rendering loss is thus defined as:
\begin{equation}
L_{\text{render}}:=\mathbb{E}_{t, \mathbf{x}_0, \epsilon}\left[\left\|\mathbf{x}_{\text{render}}-\hat{\mathbf{x}}_{\text{render}}\right\|_1\right].
\end{equation}

The final loss function for the fine stage is:

\begin{equation}
L_{\text{fine}}:=L_{\text{render}}+L_{\text{basic}}.
\end{equation}

\input{supp_sections/ddpm_alg}

\subsection{Style Guidance}
The motivation of style guidance is to address the color bias in the training dataset. 
To this end, we use the traditional clustering technique to get a style label for each $\mathbf{z}^i_0\in\mathbb{R}^{3\times4096}$ (note that $\mathbf{z}^i_0$ is defined as the color of the point cloud and not contains the coordinates). 

\paragraph{Style label.}
We commence by transforming each $\mathbf{z}^i_0$ 
in the training dataset into a feature vector. 
This transformation is realized by flattening the matrix 
into a one-dimensional array and normalizing its pixel values 
to attain zero mean and unit variance. 
The resultant feature vectors, represented as ${\mathbf{z}^i_{\text{flat}}}$, 
act as inputs for the subsequent principal component analysis (PCA). PCA is employed to reduce the dimensionality of the feature vectors. 
This process involves computing the eigenvectors 
and eigenvalues of the covariance matrix associated 
with the feature vectors. 
By retaining only the top-$n$ eigenvectors 
(where $n$ signifies the intended dimensionality of the trimmed dataset), 
we can project the feature vectors into a subspace with fewer dimensions. 
Given the global and low-frequency nature of the color style, 
we fix $n=5$ for our experiments. 
Once the reduced feature vectors are acquired, 
the K-means clustering algorithm is employed to cluster 
similarly colored point clouds. 
For our experiments, we define $k=40$.

\subsection{Hybrid Condition}

\input{supp_sections/hybrid}

\begin{figure}
    \includegraphics[trim={0cm 0cm 0cm 0cm},clip,width=\columnwidth]{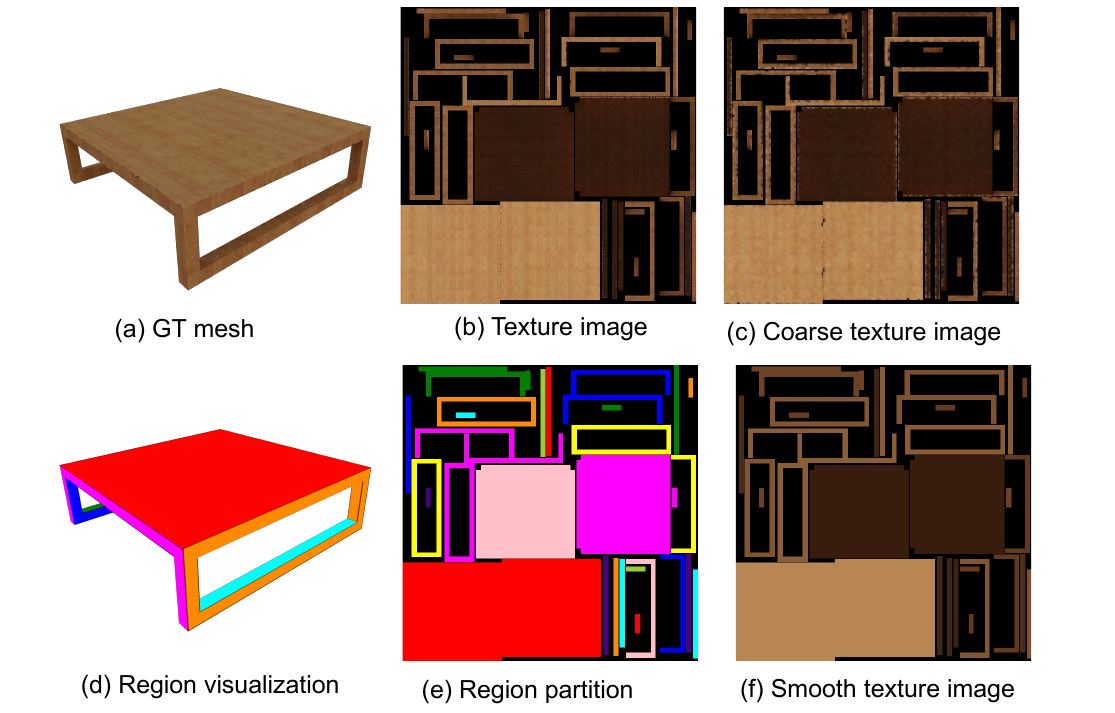}
    \caption{\textbf{Smooth map.} Different components for generating the smooth map.} 
    \label{fig:part}
\end{figure}

\paragraph{Smooth map.}
The smooth map $\mathbf{x}_{\text{smooth}}$ is obtained by smoothing the coarse UV map $\mathbf{x}_{\text{coarse}}$ by regions. Specifically, we partition the mask map into multiple distinct regions based on the detection of four-connectivity, and then apply average color pooling on each region based on its connectivity, as shown in Figure~\ref{fig:part}. After this operation, the smooth map retains only the region-based color style compared to the coarse map. 

\paragraph{Training with the hybrid condition.}
For training the fine stage, we utilize the smooth map as the primary condition and combine the original coarse map with a certain probability $p_{\text{hybrid}}$ and get a hybrid condition $\mathbf{x}_{\text{hybrid}}$. Thus, the network is forced to be capable of generating textures even when adopting a weaker condition $\mathbf{x}_{\text{smooth}}$ in the fine stage during inference. Formally, apart from the shape conditions $\mathbf{x}_{\text{shape}}$, the condition here consists of six channels, with the first three channels being the smooth map and the latter three channels set to zero values, which are randomly replaced by the coarse map with a probability of $p_{\text{hybrid}}$. The training with the hybrid condition is summarized in Algorithm~\ref{alg2}, wherein we set the $p_{\text{hybrid}}=0.3$ empirically. 

\paragraph{Denoising with condition truncation.}
After training with the hybrid condition strategy, the network is able to generate high-frequency information under weaker conditions, making it more robust to the generated results from the coarse stage during inference. 
Since the diffusion model is effective in recovering low-frequency components during the initial stages of sampling, followed by generating high-frequency components in the later stages, we employ two maps to jointly guide the generation process in the early stages and discard the coarse map in the later stages. In this way, the network will not be constrained by the inferior generated coarse map and has more flexibility for producing diversified high-frequency details.
The sampling procedure with the condition truncation is summarized in Algorithm~\ref{alg3}.

\vspace{-0.4cm}
\paragraph{Analysis.}
The hybrid condition and condition truncation are proposed for the purpose of generating high-frequency details.
The reason why the network trained and tested directly using the coarse map cannot generate texture might be due to that the coarse stage cannot be optimized perfectly, \ie, there is a domain gap between the generated coarse map and the coarse map used to train the fine stage. 
Specifically, the second stage is trained to generate high-frequency details using coarse maps with high-frequency cues, which cannot be provided by generated coarse maps from the first-stage model during testing, leading to textureless outputs.

%% file: supp_sections/ddpm_alg.tex
% \SetKwFor{DoWithNumber}{do}{until}{end do}

\begin{algorithm}
    \SetKwInOut{Input}{Input}
    \SetKwInOut{Output}{Output}
    \SetKwRepeat{Do}{do}{until}
    
    % \Input{Input data $x_1, x_2, \ldots, x_n$.}
    % \Output{Output data $y_1, y_2, \ldots, y_n$.}
    \label{alg1}
    \Do{$\text{converged}$}
    {
        $\mathbf{z}_0 \sim q(\mathbf{z}\mid \mathbf{x}_{\text{shape}}, \mathbf{z}_{\text{coord}})$\\
        $t \sim \mathcal{U}(\{1, \ldots, T\})$\\
        $\boldsymbol{\epsilon} \sim \mathcal{N}(\mathbf{0}, \mathbf{I}) $\\
        $\mathbf{z}_t=\sqrt{\bar{\alpha}_t} \mathbf{z}_0+\sqrt{1-\bar{\alpha}_t} \boldsymbol{\epsilon}$\\
        $\theta_1 \leftarrow \theta_1-\eta \nabla_{\theta_1}
        L_{\text{coarse}}\left(\mathbf{z}_t, \mathbf{z}_0, \mathbf{x}_{\text{shape}}, \mathbf{z}_{\text{coord}}\right)
        $\\
        
    } 

    \Do{$\text{converged}$}
    {
        $\left(\mathbf{x}_0, \mathbf{z}_0\right) \sim q(\mathbf{x}, \mathbf{z}\mid \mathbf{x}_{\text{shape}}, \mathbf{z}_{\text{coord}})$\\
        get $\mathbf{x}_{\text{coarse}}$ by interpolating from  $\mathbf{z}_0$\\
        $t \sim \mathcal{U}(\{1, \ldots, T\})$\\
        $\boldsymbol{\epsilon} \sim \mathcal{N}(\mathbf{0}, \mathbf{I}) $\\
        $\mathbf{x}_t=\sqrt{\bar{\alpha}_t} \mathbf{x}_0+\sqrt{1-\bar{\alpha}_t} \boldsymbol{\epsilon}$\\
        $\theta_2 \leftarrow \theta_2-\eta \nabla_{\theta_2}
        L_{\text{fine}}\left(\mathbf{x}_t, \mathbf{x}_0, \mathbf{x}_{\text{shape}}, \mathbf{x}_{\text{coarse}}\right)
        $\\
        
    } 
    \caption{Train Basic Point-UV Diffusion}
\end{algorithm}

%% file: supp_sections/hybrid.tex
% \SetKwFor{DoWithNumber}{do}{until}{end do}

\begin{algorithm}
    \SetKwInOut{Input}{Input}
    \SetKwInOut{Output}{Output}
    \SetKwRepeat{Do}{do}{until} 
    \label{alg2}
    \Do{$\text{converged}$}
    {
        $\left(\mathbf{x}_0, \mathbf{z}_0\right) \sim q(\mathbf{x}, \mathbf{z}\mid \mathbf{x}_{\text{shape}}, \mathbf{z}_{\text{coord}})$\\
        get $\mathbf{x}_{\text{coarse}}$ by interpolating from  $\mathbf{z}_0$\\
        get $\mathbf{x}_{\text{smooth}}$ by smoothing $\mathbf{x}_{\text{coarse}}$\\
        \If{$rand(0, 1) < p_{\text{hybrid}}$}
        {$\mathbf{x}_{\text{hybrid}}=\left(\mathbf{x}_{\text{smooth}}, \mathbf{x}_{\text{coarse}}\right)$}
        \Else        {$\mathbf{x}_{\text{hybrid}}=\left(\mathbf{x}_{\text{smooth}},\emptyset\right)$}
        
        $t \sim \mathcal{U}(\{1, \ldots, T\})$\\
        $\boldsymbol{\epsilon} \sim \mathcal{N}(\mathbf{0}, \mathbf{I}) $\\
        $\mathbf{x}_t=\sqrt{\bar{\alpha}_t} \mathbf{x}_0+\sqrt{1-\bar{\alpha}_t} \boldsymbol{\epsilon}$\\
        $\theta_2 \leftarrow \theta_2-\eta \nabla_{\theta_2}
        L_{\text{fine}}\left(\mathbf{x}_t, \mathbf{x}_0, \mathbf{x}_{\text{shape}}, \mathbf{x}_{\text{hybrid}}\right)
        $     
    }
    
    \caption{Training of Fine Stage with Hybrid Condition}
\end{algorithm}

\begin{algorithm}
    \SetKwInOut{Input}{Input}
    \SetKwInOut{Output}{Output}
    \SetKwRepeat{Do}{do}{until} 
    \label{alg3}
    $\mathbf{x}_T \sim \mathcal{N}(\mathbf{0}, \mathbf{I})$\\
    \For{$t=T$ to $1$}
        {
            \If{$t/T > t_c$}    {$\mathbf{x}_{\text{hybrid}}=\left(\mathbf{x}_{\text{smooth}}, \mathbf{x}_{\text{coarse}}\right)$}
            \Else        {$\mathbf{x}_{\text{hybrid}}=\left(\mathbf{x}_{\text{smooth}},\emptyset\right)$}
            $\mathbf{\hat{x}}_0 = G^2_{\theta_2}\left(\mathbf{x}_t, t, \mathbf{x}_{\text{shape}}, \mathbf{x}_{\text{hybrid}}\right)$\\
            $\mathbf{z} \sim \mathcal{N}(\mathbf{0}, \mathbf{I})$\\
            $\mathbf{x}_{t-1}=\frac{\sqrt{\bar{\alpha}_{t-1}} \beta_t}{1-\bar{\alpha}_t} \mathbf{\hat{x}}_0+\frac{\sqrt{\alpha_t}\left(1-\bar{\alpha}_{t-1}\right)}{1-\bar{\alpha}_t} \mathbf{x}_t
            +\sigma_t \mathbf{z}$ 
        
        }
    \Return $\mathbf{\hat{x}}_0$

    \caption{Sampling of Fine Stage with Condition Truncation}
\end{algorithm}

%% file: supp_sections/implementations.tex
\section{Implementation Details}
\label{sec:sup_implementation}
\subsection{UV Map Process}

\begin{figure}[t]
\raggedright
    \includegraphics[trim={0cm 0cm 0cm 0cm},clip,width=\columnwidth]{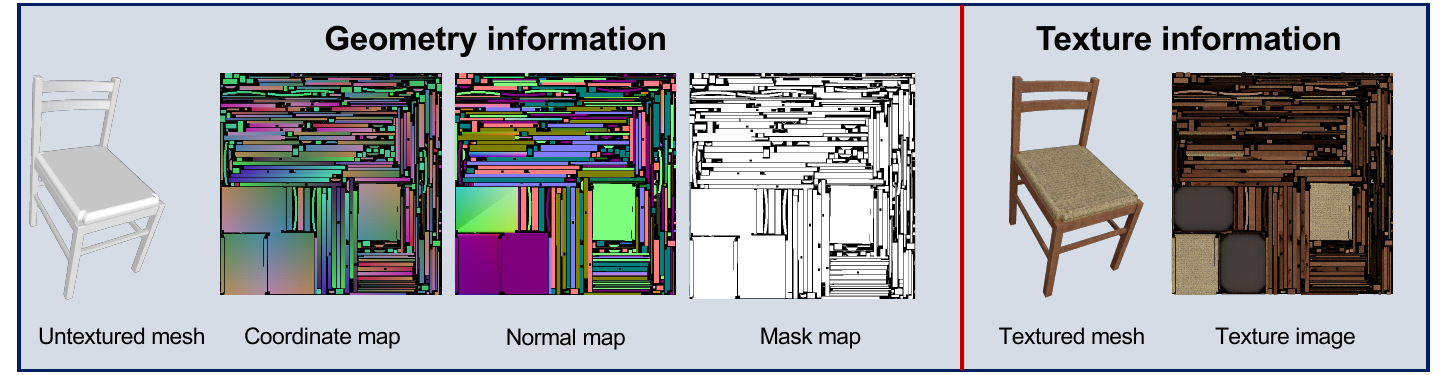}
    \caption{\textbf{Different information encoded in UV space.}}
    \label{fig:notion}
    \vspace{-0.4cm}
\end{figure}

\begin{table*}[htbp]
\centering
\begin{tabular}{|c|c|cccc|}
\hline Level & Blocks Type & Number of Blocks & Block Channel & Dropout & Self Attention \\
\hline 1 & Encoder & 1 & 32 & 0.0 & False \\
2 & Encoder & 2 & 64 & 0.0 & False \\
3 & Encoder & 2 & 128 & 0.0 & False \\
4 & Encoder & 2 & 192 & 0.0 & False \\
5 & Encoder & 4 & 384 & 0.2 & True \\
\hline 5 & Decoder & 5 & 384 & 0.2 & True \\
4 & Decoder & 3 & 192 & 0.0 & False \\
3 & Decoder & 3 & 128 & 0.0 & False \\
2 & Decoder & 3 & 64 & 0.0 & False \\
1 & Decoder & 2 & 32 & 0.0 & False \\
\hline
\end{tabular}
\vspace{2pt}
\caption{\textbf{Details of our U-Net network.}}
\label{tab:network}
\end{table*}
\vspace{-0.5pt}

We now describe how we process the raw data from ShapeNet to obtain our data format. Firstly, we utilize xatlas~\cite{JonathanYoung18} to automatically generate UV mapping for each mesh. This allows us to establish a correspondence between the surface points and 2D planes, denoted by function $f$. Subsequently, we select a resolution $H\times W$ for the 2D plane. In our case, we use a resolution of 512$\times$512. With this resolution, we discretize the 2D plane to obtain the coordinates of each pixel in UV space. As shown in Figure~\ref{fig:notion}, by applying function $f$, we are able to calculate the corresponding 3D coordinates for each pixel, which results in a coordinate map $\mathbf{x}_{\text{coord}}\in\mathbb{R}^{3\times H\times W}$. It is worth noting that due to topological reasons, some pixels do not have corresponding 3D coordinates. This occurs when there are no surface points that are mapped to these pixels. These pixels are marked as invalid and we can obtain a binary mask map, as shown in Figure~\ref{fig:notion}, to which we refer to the mask map. Similarly, we can compute the surface normal for each point and record them in their corresponding UV coordinates and get a normal map. Note that, these information only depends on the geometry of the mesh, which we use as shape conditions in our method.

To obtain the texture image, we use Blender to render each mesh from multiple viewpoints (\ie, 50 views), and then we back-project the resulting multi-view images into 3D space using the camera matrices to obtain a corresponding dense colored point cloud $P_{src}$. For creating the ground-truth texture image with RGB colors, we query the 3D coordinates of each pixel and use KNN (in our case, we use 3-nearest neighbors) interpolation to calculate its corresponding color from $P_{src}$.

\subsection{Training Configurations}
We implement all the experiments using PyTorch on four NVIDIA RTX 3090 GPU cards with batch size 8. The learning rate is initially set to 0.0002 and scheduled by cyclic cosine annealing~\cite{loshchilov2016sgdr}, and models are optimized by AdamW~\cite{loshchilov2017decoupled} with $\beta_1$ = 0.9 and $\beta_2$ = 0.999. For the coarse stage, we train 500 epochs for bench category and 400 epochs for other categories. For the fine stage, we train 150 epochs for the Chair and Car categories and 250 epochs for other categories. The ablation studies are conducted in the same settings. We use an EMA rate of 0.9995 for all experiments. For training PVD-Tex, we sample a point cloud with 512$\times$512 points on the surface, where each point corresponds to a pixel of the UV map with resolution 512$\times$512, and train a diffusion network to learn the RGB value for each point using the same network architecture and training scheme as in~\cite{zhou20213d}. Then, we map the corresponding point colors onto the UV map, such that we can derive the textured mesh with this UV map.

\subsection{Network Architecture}

\begin{figure}
    \includegraphics[trim={0cm 0cm 0cm 0cm},clip,width=\columnwidth]{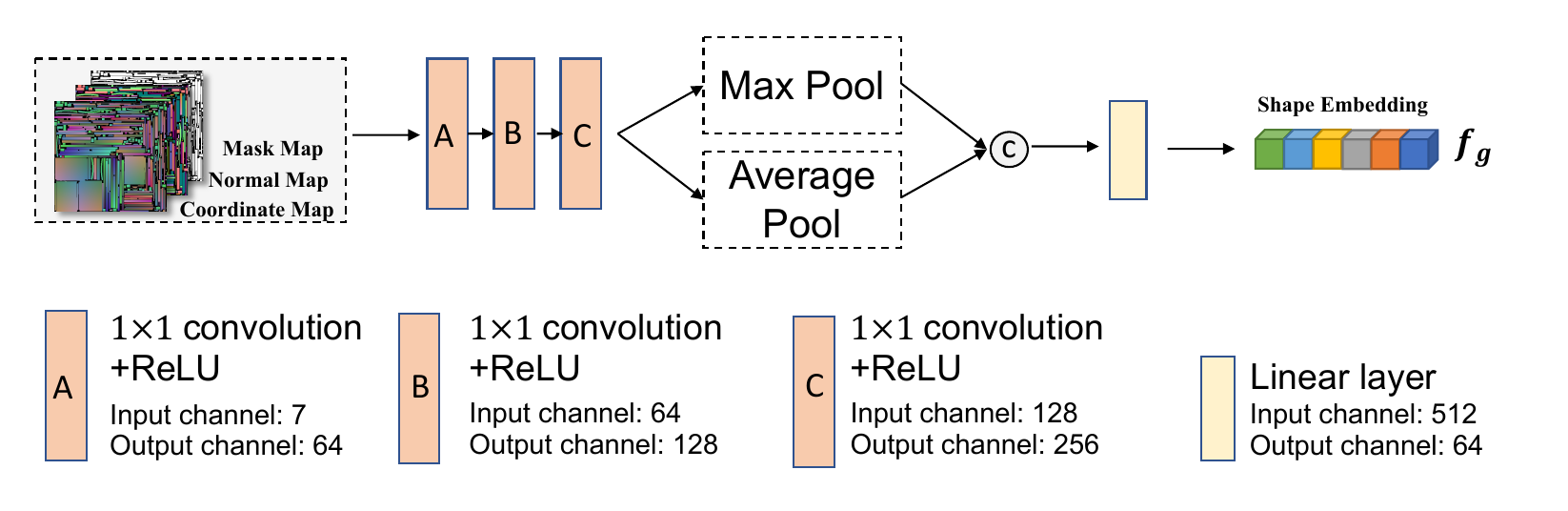}
    \caption{\textbf{Details of our shape encoder.} }
    \label{fig:encoder}
\end{figure}

\paragraph{Coarse stage.}
The configurations of the network in our coarse stage are based on those in PVD~\cite{zhou20213d}, except for some modifications. In particular, PVD has two types of inputs, \ie noisy point cloud coordinates and time step. In contrast, we have five types of inputs, noisy point cloud colors $\mathbf{z}_t\in\mathbb{R}^{3\times 4096}$, clean point cloud coordinates $\mathbf{z}_{\text{coord}}\in\mathbb{R}^{6\times 4096}$(attached with its normal), shape map $\mathbf{x}_{\text{shape}}\in\mathbb{R}^{7\times 512\times512}$, style label $z_{\text{style}}$ and time step $t$. To align with PVD, we first use a shape encoder $E_{\phi}$ to extract a global feature $f_g\in\mathbb{R}^{64\times 1}$ from $\mathbf{x}_{\text{shape}}$. 
Then, we expand $f_g$  in the last dimension to align with the point number of $\mathbf{z}_t$, \ie we get $f^{\text{expand}}_g\in\mathbb{R}^{64\times 4096}$. Next, we concatenate $f^{\text{expand}}_g, \mathbf{z}_t, \mathbf{z}_{\text{coord}}$ through the channel dimension,  thereby aligning the point cloud input in PVD. To handle $z_{\text{style}}$, note that the network uses \textit{nn.embedding} and linear layer to extract a time embedding from timestep $t$. Therefore, we perform the same operations on $z_{\text{style}}$ and add its extracted embedding to the time embedding before feeding it into the network.
The architecture of $E_{\phi}$ is shown in Figure~\ref{fig:encoder}.

\vspace{-0.5cm}
\paragraph{Fine stage.}
We build our fine stage on a 2D U-Net structure based on guided diffusion~\cite{dhariwal2021diffusion}. In our case, we concatenate the coarse map, smooth map, shape map, and the noisy UV texture image along the channel dimension to obtain the input tensor $\mathbf{x}_{\text{input}}\in\mathbb{R}^{16\times512\times512}$.
We utilize five levels for the U-Net architecture in both the encoder and decoder, incorporating dropout and self-attention at the lowest-resolution level. Each level's basic block employs the residual block built from~\cite{dhariwal2021diffusion}, with the number of blocks and convolutional channel numbers specified in Table~\ref{tab:network}.